\documentclass[11pt]{article}

\usepackage[final]{acl}

\usepackage{booktabs}
\usepackage{multirow}
\usepackage{bbding}
\usepackage{diagbox}
\usepackage{times}
\usepackage{latexsym}
\usepackage{amssymb}
\usepackage{amsmath}
\usepackage{siunitx}
\usepackage{tcolorbox}
\usepackage{xcolor}

\newcommand{\myaclprompt}[1]{
	\begin{center}
		\begin{tcolorbox}[
			colback=black!85,
			colframe=black!85,
			arc=1mm,
			boxrule=0pt,
			width=0.95\linewidth, % 强制与行宽一致
			sharp corners=south, % 底部直角，方便对齐
			halign=center,
			valign=center,
			fontupper=\small\bfseries\sffamily\color{white},
			height=1.5em % 模拟标题栏高度
			]
			P(True) Prompt
		\end{tcolorbox}
		
		\vspace{-0.7cm} % 关键：微调缝隙，让两个框视觉上“咬合”在一起
		
		\begin{tcolorbox}[
			colback=gray!10,
			colframe=black!85, % 浅灰色下框
			arc=1mm,
			boxrule=1.2pt,
			width=0.95\linewidth,
			sharp corners=north, % 顶部直角，方便对齐
			top=10pt,
			left=12pt,
			right=12pt,
			bottom=10pt
			]
			{\small #1} % 这里放置内容
		\end{tcolorbox}
	\end{center}
}

\usepackage[T1]{fontenc}
\usepackage[utf8]{inputenc}

\usepackage{microtype}

\usepackage{inconsolata}

\usepackage{graphicx}

\title{CausalGaze: Unveiling Hallucinations via Counterfactual Graph Intervention in Large Language Models}

\author{Linggang Kong$^\text{1,2}$, Lei Wu$^\text{1,2}$, Yunlong Zhang$^\text{1,2}$, Xiaofeng Zhong$^\text{1,2}$ \\\textbf{Zhen Wang$^\text{1,2}$, Yongjie Wang$^\text{1,2}$\thanks{Corresponding author}, Yao Pan$^\text{3}$}\\  \\
	{$^\text{1}$College of Electronic Engineering, National University of Defense Technology}\\
	{$^\text{2}$Anhui Province Key Laboratory of Cyberspace Security Situation Awareness and Evaluation}\\
	{$^\text{3}$Institute of Computer Application, China Academy of Engineering Physics}\\
	\texttt{\{konglinggang,wangyongjie17\}@nudt.edu.cn}}

\begin{document}
\maketitle
\begin{abstract}
Despite the groundbreaking advancements made by large language models (LLMs), hallucination remains a critical bottleneck for their deployment in high-stakes domains. Existing classification-based methods mainly rely on static and passive signals from internal states, which often captures the noise and spurious correlations, while overlooking the underlying causal mechanisms. To address this limitation, we shift the paradigm from passive observation to active intervention by introducing CausalGaze, a novel hallucination detection framework based on structural causal models (SCMs). CausalGaze models LLMs' internal states as dynamic causal graphs and employs counterfactual interventions to disentangle causal reasoning paths from incidental noise, thereby enhancing model interpretability. Extensive experiments across four datasets and three widely used LLMs demonstrate the effectiveness of CausalGaze, especially achieving 3.3\% AUROC improvement in AUROC on the TruthfulQA dataset compared to state-of-the-art baselines.
\end{abstract}

\section{Introduction}

Large language models (LLMs) excel at various natural language generation and reasoning tasks, yet hallucination, where models generate plausible but factually incorrect content, remains a significant barrier for real-world deployment \citep{bib1}. The prevalence of hallucination fundamentally undermines the trustworthiness and reliability of LLM-based systems, necessitating effective and robust hallucination detection and mitigation mechanisms \citep{bib2,bib3,zhang2026stable}. Efforts to detect hallucinations mainly include retrieval-based methods \citep{bib4,bib5}, which require searching external knowledge bases, while consistency-based and logits-based methods often perform multiple inferences for consistency and entropy calculations \citep{bib6,bib7}. In contrast, classification-based methods utilizes the semantic features of internal states, requiring only a single generation pass without relying on any external sources \citep{bib8}.

\begin{figure}[t]
	\includegraphics[width=\columnwidth]{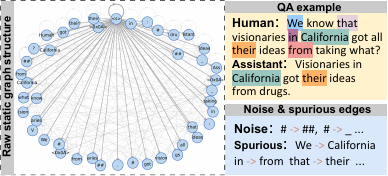}
	\caption{An question-answering example with noise and spurious correlations in graph structure. The nodes and edges are from the raw hidden states and attention maps, respectively.}
	\label{example}
\end{figure}

Existing hallucination detection methods based on internal states typically train the classifiers either on hidden states or attention maps \citep{bib9,Lookback}. And these methods have demonstrated effective performance across various datasets and model architectures, as shown in multiple studies \citep{bib10,bib11,bib12,bib13}. To jointly capture the semantic features and token dependencies, recent work has explored static graph structures \citep{bib14}, where the hidden states and attention maps serve as the nodes and edges, respectively. While the graph-based classifier has superior performance, it is inherently susceptible to capturing noise and spurious correlations from the raw graph structure as shown in Figure~\ref{example}. Such vulnerabilities can lead the classifier to learn and propagate incorrect dependencies. Furthermore, the utility and generalization of the hallucination detector are significantly undermined by the indiscriminate aggregation of information without a causal basis. The fundamental challenge lies in that this passive observation paradigm fails to differentiate structurally robust knowledge from fragile associative patterns \citep{bib15}.

To overcome these limitations, we propose a paradigm shift from passive observation to active intervention to accurately trace the causal path between the internal information and model outputs. We posit that factual content is structurally robust, whereas hallucination stems from noise associations, thus highly sensitive to micro-structural interventions. Based on this insight, we propose CausalGaze, a novel hallucination detection framework based on structural causal models (SCMs) \citep{bib16}. Specifically, we model the LLMs' hidden states and attention maps as dynamic causal graphs and employ gradient-guided counterfactual intervention in the raw graph structure. To the best of our knowledge,  CausalGaze is the first work to introduce active causal intervention to address the passive attribution in hallucination detection tasks, offering a novel solution with causal interpretability in this field. Our main contributions are summarized as follows:
\begin{itemize}
	\setlength{\itemsep}{0pt} % 设置间距为0pt
	\item We propose CausalGaze, a novel hallucination detection framework that models internal states as dynamic causal graphs. We are the first to use the gradient-guided counterfactual intervention to estimate the causal sensitivity of attention edges ($\nabla A$), disentangling causal dependencies from spurious connections.
	\item We introduce a method to derive interpretable causal subgraphs by integrating node gradients ($\nabla H$) with the causally refined edges, providing fine-grained insights into the causal origins and paths that lead to hallucinations.
	\item We evaluate CausalGaze across four datasets and three widely used LLMs and compare the performance with baseline methods. The results demonstrate the significant effectiveness of CausalGaze, achieving over 3.3\% improvement on the TruthfulQA dataset.
\end{itemize}

\section{Related Work}

\textbf{Hallucination Detection.} The remarkable generative capacity of Large Language Models (LLMs) has enabled their widespread application in knowledge-intensive and reasoning tasks \citep{bib1}. However, LLM hallucination remains a critical bottleneck, hindering their deployment in high-stakes domains (e.g., healthcare \citep{bib17}, finance \citep{bib18}, and cybersecurity). Hallucination severely compromises model trustworthiness, motivating extensive efforts to detect it \citep{bib19}. Existing methods for hallucination detection are primarily categorized into two types. Black-box approaches rely on external knowledge checking (e.g., search engines or RAG-based verification) \citep{bib4,add4} or consistency checks (e.g., self-consistency) \citep{bib20}. While these methods are easy to implement, their effectiveness is limited by the correctness and completeness of external knowledge sources and fail to provide root-cause analysis for model internal errors \citep{bib6}. In contrast, white-box methods aim to provide deeper and sourced insights by analyzing the LLMs’ internal states. Prior work has mainly focused on the model’s logical outputs \citep{bib21} or latent space, employing techniques such as entropy calculation \citep{bib22,bib7}, feature extraction \citep{bib23}, clustering \citep{bib12}, and classifier training \citep{bib24,bib10}. Critically, these methods often capture noise and spurious correlations in the raw information, and only identify superficial correlational relationships. Therefore, we are the first to detect hallucination by performing active interventions from the causal perspective.

\begin{figure*}[t]
	\centering 
	\includegraphics[width=0.95\textwidth]{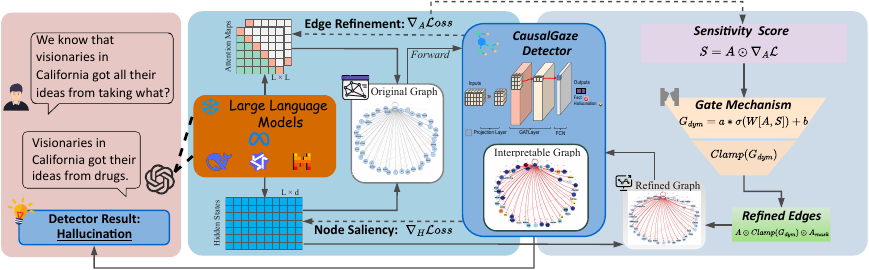}
	\caption{Overall framework of the proposed CausalGaze. We first employ the gradient-guided counterfactual intervention to obtain the refined edges, then the hidden states and the refined edges are together passed to the GNN-based detection module for the final hallucination detection result.}
	\label{frame}
\end{figure*}

\textbf{Graph Causal Learning.} The interpretability of Graph Neural Network (GNN), such as GNNExplainer \citep{bib25}, PGExplainer \citep{bib26}, typically uses techniques like gradients or perturbations to identify the most influential subgraphs for the prediction. However, these tools are inherently post-hoc and attributional \citep{bib14}. The primary goal is to explain why the model predicts, rather than serving as an intrinsic optimization mechanism to correct how the model reasons and thinks. We draw upon the concept of gradient sensitivity to estimate causality but repurpose it to drive a learnable structural intervention module. Furthermore, integrating it directly into the framework is a key direction for enhancing model robustness. The applications typically focus on mitigating confounding and selection biases in graph datas with fixed and real-world topologies (e.g., social or knowledge graphs) \citep{bib27}. In this paper, we extend this paradigm to a novel domain: graph-structure information generated by LLMs, and offer an innovative and interpretable solution for LLM hallucination detection tasks.

\section{Methodology}
\label{sec3}
%% Labels are used to cross-reference an item using \ref command.

The previous work \citep{bib14} has found that weighted directed graphs $G=(V,E)$ can effectively integrate token semantic information and their dependencies, where the nodes $V$ are defined by the sequence of hidden states $H \in \mathbb{R}^{L \times d}$ ($L$ is the sequence length, $d$ is the dimension), and the edges $E$ are represented by the attention maps $A \in \mathbb{R}^{L \times L}$. The detection model learns the correlation between the graph $G$ and the label $Y$ to predict $P(Y|G)$ with competitive performance. Nevertheless, it is hindered because the raw attention maps $A$ contain massive unrelated connections, fostering illusory learning patterns with noise.

To disentangle these factors, we formulate a structural causal model (SCM) represented by the directed acyclic graph: $C \to A \to Y$ and $C \to Y$. Where $C$ denotes the confounder, $A$ is the mediator, i.e., the attention maps, $Y$ is the label. Our objective is to estimate the causal effect of the mediator on the prediction, denoted as $P(Y | H,do(A))$, rather than the mere observational connections.
\subsection{Problem setup}
\label{subsec3.1}
Following the core assumption of SCM, our central hypothesis is that factual reliability corresponds to structural robustness. Conversely, hallucinations are rooted in fragile and irrelevant associations. Therefore, we transform the static weighted directed graph $G$ into a dynamic causal graph $\mathcal{G} = (\mathcal{V}, \mathcal{E}_{causal})$. The nodes $\mathcal{V}$ are still defined by the sequence of hidden states $H$, while the edges $\mathcal{E}_{causal}$ are denoted by the causal attention maps $\tilde{\mathbf{A}}$. And the hallucination detection task is framed as a binary classification problem, and the goal is to learn a causal mapping $f$, as shown in Equation~\ref{eq1}:
\begin{equation}
	\label{eq1}
	y=f(\mathcal{G}), \,\,\,y \in \{0, 1\}
\end{equation}
where $y$ denotes the predicted label, $y=0$ represents `fact', and $y=1$ represents `hallucination'.
\subsection{CausalGaze Framework}

\label{subsec3.2}
The proposed CausalGaze framework achieves the paradigm shift from passive observation to active intervention through a dual-mechanism pipeline as illustrated in Figure~\ref{frame}.

\textbf{Edge Refinement}: We employ the gradient-guided counterfactual intervention to compute the causal sensitivity of each edge $A_{ij}$, which drives a learnable gating mechanism to generate a refined and causal edges $\tilde{\mathbf{A}}$. The hidden states $H$ and the refined edges $\tilde{\mathbf{A}}$ are then passed to a downstream GNN-based detection module $D(\cdot)$ for the final classification.
 
\textbf{Node Saliency}: We utilize the node gradient ($\nabla H$) as a complementary mechanism to identify the most salient tokens contributing to the final decision. The most salient tokens and refined edges are jointly to obtain the token-level interpretable causal subgraphs.

\subsubsection{Gradient-Guided Counterfactual Intervention}
%% Inline mathematics is tagged between $ symbols.
The core of CausalGaze is the counterfactual intervention mechanism, which actively disentangles semantic edges from noise and spuriousness. However, calculating the exact causal effect via $do$-calculus requires traversing all possible confounders, which is computationally intractable in high-dimensional model spaces. Therefore, we propose to use the gradient-guided counterfactual intervention approximation. We define the causal sensitivity of an edge $(j \to i)$ as the magnitude of change in the loss $\mathcal{L}_{Detector}$ under a microscopic intervention on the edge weight $A_{ij}$. This corresponds to the individual treatment effect (ITE) in a local neighborhood, which could be considered as a perturbation $\epsilon$ on the weight of edge $A_{ij}$. By applying the first-order $Taylor$ expansion, the change in the loss $\mathcal{L}_{Detector}$ of the hallucination detector can be approximated, as shown in Equation~\ref{eq2}:
\begin{equation}
	\label{eq2}
	\mathcal{L}(A_{ij} + \epsilon) - \mathcal{L}({A}_{ij}) \approx \epsilon \cdot \frac{\partial \mathcal{L}_{Detector}}{\partial {A}_{ij}}
\end{equation}

The gradient $\frac{\partial \mathcal{L}_{Detector}}{\partial {A}_{ij}}$ indicates the direction of steepest descent, which does not account for the magnitude of the information flow, as dipicted in Figure~\ref{gradient}. Thus, we define the causal sensitivity matrix ${S} \in \mathbb{R}^{N \times N}$ via the hadamard product of the edge weights and their gradients:

\begin{figure}[t]
	\centering 
	\includegraphics[width=\columnwidth]{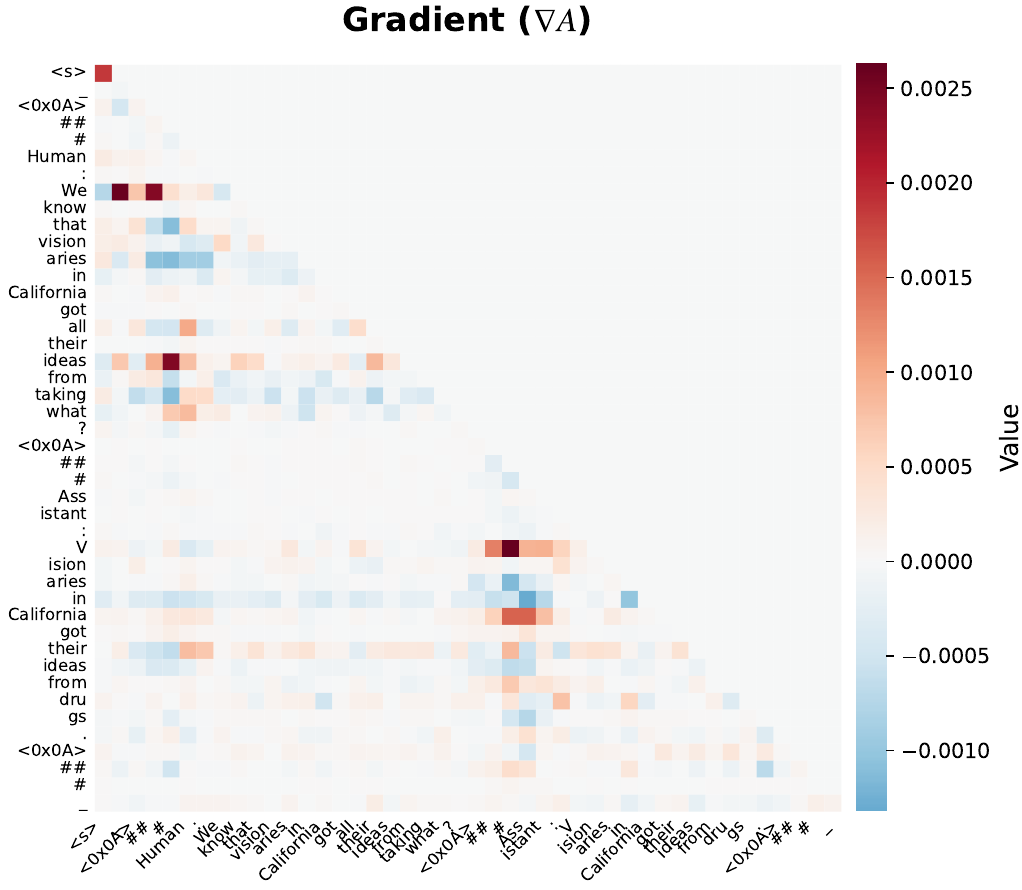}
	\caption{Visualization of the gradients for the example in Figure~\ref{example}.The noise connection from token `\textit{\#}' to token `\textit{\#\#}' and spurious connection from token `\textit{We}' to token \textit{`California}' have a near-zero value.}
	\label{gradient}
\end{figure}

\begin{equation}
	\label{eq3}
	{S} = \left| {A} \odot \nabla_{{A}}\mathcal{L}_{Detector} \right|
\end{equation}
where $\odot$ denotes element-wise multiplication. A spurious or noise edge might have a high weight due to position bias, but a near-zero gradient for irrelevance to reasoning. Conversely, a causal and factual edge should possess both significant weight and high gradient sensitivity.

\begin{figure}[t]
	\centering
	% 第一张图
	\begin{minipage}[b]{0.49\columnwidth}
		\centering
		\includegraphics[width=\textwidth]{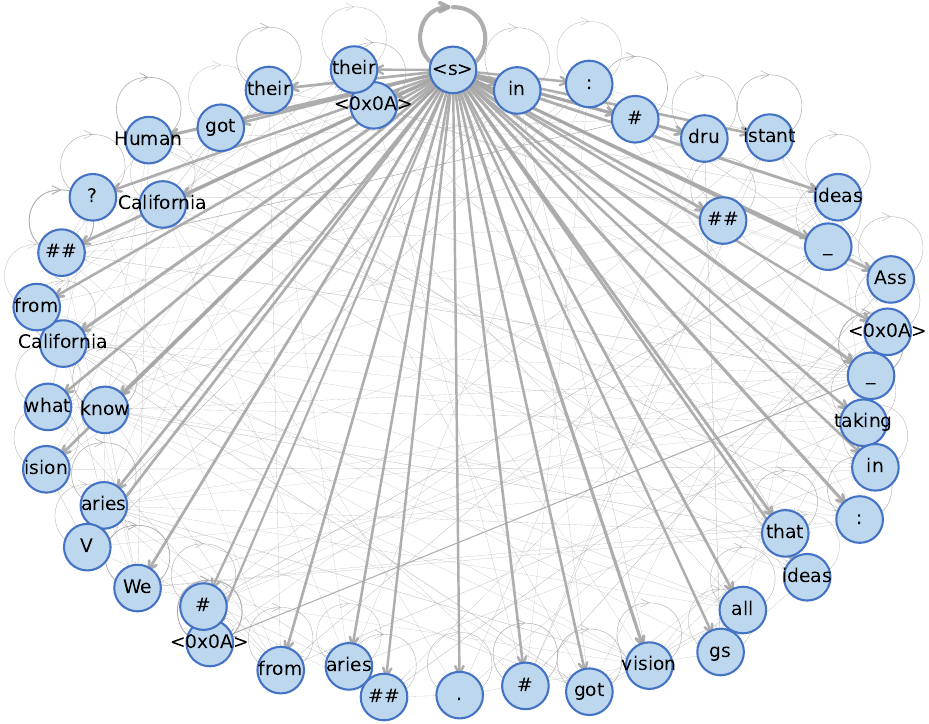}
		\caption*{(a) Raw graph} % 如果不需要子标题，可以删掉
	\end{minipage}
	\hfill % 在两图之间插入弹性间距
	% 第二张图
	\begin{minipage}[b]{0.49\columnwidth}
		\centering
		\includegraphics[width=\textwidth]{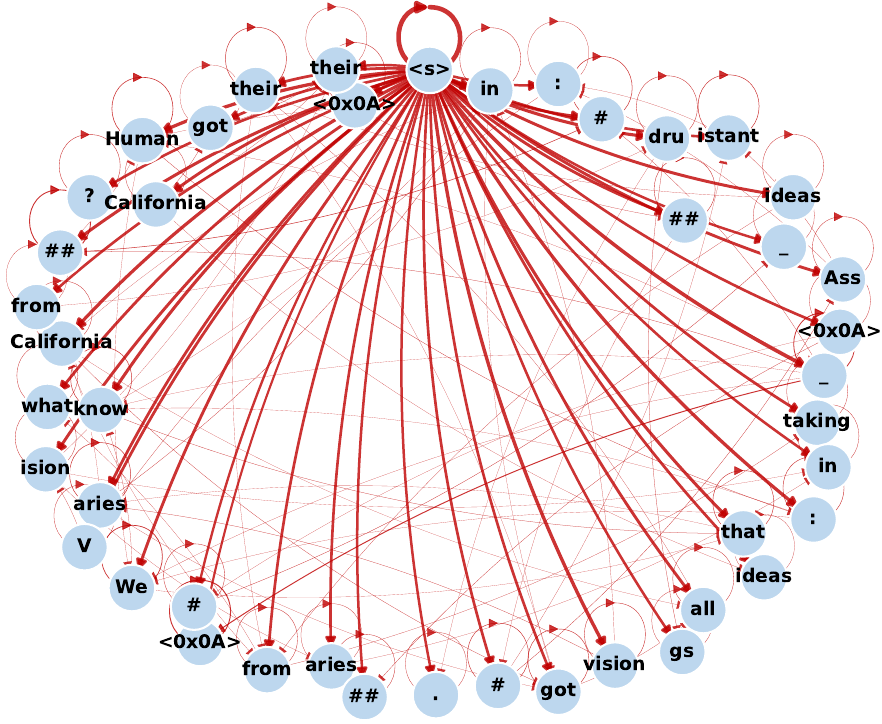}
		\caption*{(b) Refined graph}
	\end{minipage}
	
	\caption{The comparison of graph structures before and after the intervention for the example in Figure~\ref{example}. (a) The edges in raw graph are from the observation attention maps. (b) The edges in refined graph are from the actively intervention attention maps.}
	\label{refined}
\end{figure}

To transform the raw sensitivity $S$ into actively intervened edges $\tilde{\mathbf{A}}$, we introduce a learnable causal refinement layer (CRL). The CRL employs a dynamic gating mechanism that not only suppresses noise but also amplifies crucial, yet potentially weak and causal links. The dynamic gate mechanism is a shallow network MLP with both the original attention maps $A$ and causal sensitivity $S$ as input:
\begin{equation}
	\label{eq4}
	G_{dym} =a* \boldsymbol{\sigma} (MLP([A,S])) + b
\end{equation}
where $[\cdot,\cdot]$ is the concatenation operation, $\boldsymbol{\sigma}$ denotes the sigmoid function and $a,b$ are the learnable scaling factors. The final refined edges $\tilde{\mathbf{A}}$ is obtained as follows:
\begin{equation}
	\label{eq5}
	\tilde{\mathbf{A}} = {A} \odot Clamp(G_{dym}) \odot {A}_{mask}
\end{equation}
where $Clamp(\cdot)$ is the clamp function to further enhance the intervention and ${A}_{mask}$ is the autoregressive mask of ${A}$. The actively intervened token dependencies $\tilde{\mathbf{A}}$ mitigates noise and spurious connections in the raw information. And the comparison of graph structures before and after the intervention is shown in Figure~\ref{refined}, revealing that the refined graph is more sparsely connected.

\subsubsection{CausalGaze Detector}
%% Inline mathematics is tagged between $ symbols.
The Graph Attention Network (GAT) backbone is deployed on the refined causal graph $\tilde{\mathcal{G}}= (\mathcal{V}, \mathcal{E}_{causal})$ to aggregate information, which integrates the refined edges as explicit structural bias. The update of node $i$  at layer $k$ is given in Equation~\ref{eq6}:
\begin{equation}
	\label{eq6}
	\mathbf{h}_i^{(k+1)} =  \mathbf{h}_i^{(k)} + \sum_{j \in \mathcal{N}(i)} \tilde{\mathbf{A}}^{(k)}_{ij} \mathbf{W}^{(k)} \mathbf{h}_j^{(k)}
\end{equation}
where $\mathbf{W}^{(k)}$ is the learnable parameters and multiple GAT layers with residual connections is stacked to facilitate deep message passing. To capture both the most salient local features and the global context, we employ a hybrid pooling strategy that concatenates the results of global max pooling and mean pooling:
\begin{equation}
	\label{eq7}
	\mathbf{h}_{graph} = \left[ \max_{i \in \mathcal{V}} \mathbf{h}_i^{(K)} , \frac{1}{|\mathcal{V}|} \sum_{i \in \mathcal{V}} \mathbf{h}_i^{(K)} \right]
\end{equation}

The final representation $\mathbf{h}_{graph}$ is input into an MLP layer for classification. The objective combines the standard binary classification loss with a regularization term that promotes sparsity and coherence in the causal structure:
\begin{equation}
	\label{eq8}
	\mathcal{L}_{Detector} = \mathcal{L}_{CE}(y, \hat{y}) + \lambda \cdot \|{S}\|_F^2
\end{equation}
where $\mathcal{L}_{CE}$ is the primary cross-entropy loss between the true label $y$ and the prediction $\hat{y} = f(\mathcal{G})$,  $\lambda$ is applied to the sensitivity matrix $S$ to encourage the model to find the most compact, sparse set of causally important edges.
\subsection{Interpretable causal subgraphs}
\label{subsec3.3}
To further elaborate the interpretability of hallucination detection model CausalGaze, we complement the structural analysis with node-level attribution. We utilize the gradient of the hidden states ($\nabla H$) as a diagnostic lens to calculate the importance scores of each token without altering the semantic features. The node saliency score $N^i_{s}$ for the $i\mbox{-}th$ token is calculated as the $L_2$ norm of the gradient vectors:
\begin{equation}
	\label{eq9}
	N^i_{s} = \| \nabla_{H_i} \mathcal{L}_{Detector} \|_2, \,\,\, i\in[1,L]
\end{equation}
where $H_i$ denotes the hidden states of the $i\mbox{-}th$ token. The interpretable causal subgraphs could be obtained by combining the refined edges $\tilde{\mathbf{A}}$ and the salient nodes $\tilde{\mathbf{N}}$, which allows for a fine-grained diagnosis to illustrate the causal nodes and paths from the token-level perspective.

\section{Experimental Settings}
\label{sec4}

\subsection{Models}
\label{subsec4.1}
We evaluate the detection performance of CausalGaze using three mainstream open-source LLMs, including Llama-2-7B \citep{add1}, Qwen2-7B \citep{add2}, and Mistral-7B \citep{add3}. Details of  the settings of these models are in Appendix~\ref{appendixA1}

\subsection{Datasets}
\label{subsec4.2}
The models are evaluated on four widely used datasets:  TruthfulQA \citep{bib28}, which is a open-domain question answering dataset; TriviaQA \citep{bib29}, which tests general knowledge; SicQ \citep{bib30}, designed for domain-specific question answering; and  HaluEval \citep{bib31}, specially designed for hallucination detection tasks.
\subsection{Baselines}
\label{subsec4.3}
We compare our approach against several types of competitive baselines, categorized as follows: (1) \textbf{Consistency-based} approaches: SelfcheckGPT (black-box) \citep{bib6} and EigenScore (white-box) \citep{bib8}; (2) \textbf{Logit-based} approaches: Perplexity \citep{add6}, Length-Normalized Entropy (LN-Entropy) \citep{add5}, and Semantic Entropy \citep{bib22}; (3) \textbf{Self-evaluation} approach: P(True) \citep{add7}; and (4) \textbf{Classification-based} approaches: SAPLMA \citep{bib24}, LLM-Check \citep{bib12}, ICR Probe \citep{bib13} and HaluGNN \citep{bib14}. All comparison methods were implemented using the experimental parameters specified in their respective original papers. More details are shown in Appendix~\ref{appendixA3}.
\subsection{Evaluation Metric}
\label{subsec4.4}
Following previous established research, we employ the Area Under the Receiver Operating Characteristic Curve (AUROC) as the primary evaluation metric to assess the performance of all approaches \citep{bib24, bib12, bib13}. AUROC represents the area under the ROC curve, which illustrates the trade-off between the True Positive Rate (TPR) and the False Positive Rate (FPR). Furthermore, the F1-Score is also selected as an essential evaluation metric, offering a balanced measure of the model's performance on classification tasks.

% Please add the following required packages to your document preamble:
% \usepackage{multirow}
\begin{table*}[t]
	\centering
	
	\resizebox{\textwidth}{!}{
		\begin{tabular}{clcccccccc}
			\toprule
			\multirow{2}{*}{\textbf{LLMs}}        & \multicolumn{1}{c}{\multirow{2}{*}{\textbf{Methods}}} & \multicolumn{2}{c}{\textbf{TruthfulQA}}    & \multicolumn{2}{c}{\textbf{TriviaQA}}      & \multicolumn{2}{c}{\textbf{SciQ}}          & \multicolumn{2}{c}{\textbf{HaluEval}}      \\ \cline{3-10} 
			& \multicolumn{1}{c}{}                         & AUROC           & F1-Score        & AUROC           & F1-Score        & AUROC           & F1-Score        & AUROC           & F1-Score        \\ \midrule
			\multirow{11}{*}{Llama2-7B}  & SelfCheckGPT                                 & 0.5295          & 0.5345          & 0.7322          & 0.7296          & 0.6790          & 0.6830          & 0.6670          & 0.6591          \\
			& EigenScore                                   & 0.5193          & 0.5286          & 0.7398          & 0.7443          & 0.5960          & 0.6008          & 0.5960          & 0.5880          \\
			& Perplexity                                   & 0.5677          & 0.6034          & 0.7213          & 0.6995          & 0.5260          & 0.5256          & 0.5260          & 0.5020          \\
			& LN-Entropy                                   & 0.6151          & 0.6187          & 0.7091          & 0.7332          & 0.5760          & 0.5826          & 0.6678          & 0.6544          \\
			& Semantic Entropy                             & 0.6217          & 0.6145          & 0.7321          & 0.7209          & 0.6820          & 0.6797          & 0.6820          & 0.6865          \\
			& P(True)                                      & 0.5181          & 0.5556          & 0.5568          & 0.5704          & 0.5460          & 0.5645          & 0.5648          & 0.6150          \\
			& SAPLMA                                       & 0.7820          & 0.7903          & 0.8310          & 0.8520          & 0.7730          & 0.7767          & 0.7738          & 0.6936          \\
			& LLM-Check                                    & 0.6160          & 0.5926          & 0.5551          & 0.5857          & 0.5842          & 0.5637          & 0.5683          & 0.5450          \\
			& ICR-Probe                                    & 0.8142          & 0.7858          & 0.8001          & \textbf{0.7940} & 0.7748          & 0.7561          & 0.8346          & \textbf{0.8032} \\
			& HaluGNN                                      & 0.8803          & 0.7714          & 0.8531          & 0.7863          & 0.8336          & 0.7649          & 0.9141          & 0.6316          \\
			& \textbf{CausalGaze}                          & \textbf{0.8805±0.0102} & \textbf{0.8189±0.0510} & \textbf{0.8614±0.0110} & 0.7786±0.0211          & \textbf{0.8390±0.0081} & \textbf{0.7701±0.0420} & \textbf{0.9280±0.0204} & 0.8001±0.0210          \\ \midrule
			\multirow{11}{*}{Qwen2-7B}   & SelfCheckGPT                                 & 0.6170          & 0.6220          & 0.6230          & 0.6143          & 0.5860          & 0.5779          & 0.6538          & 0.6327          \\
			& EigenScore                                   & 0.5370          & 0.5489          & 0.6130          & 0.6180          & 0.6320          & 0.6462          & 0.6840          & 0.6343          \\
			& Perplexity                                   & 0.6510          & 0.6630          & 0.5020          & 0.5103          & 0.5340          & 0.5477          & 0.5340          & 0.5167          \\
			& LN-Entropy                                   & 0.6670          & 0.6594          & 0.5110          & 0.5063          & 0.5240          & 0.5256          & 0.7371          & 0.6850          \\
			& Semantic Entropy                             & 0.6610          & 0.6787          & 0.5870          & 0.5965          & 0.6590          & 0.6604          & 0.6634          & 0.6808          \\
			& P(True)                                      & 0.6370          & 0.6542          & 0.5090          & 0.5256          & 0.5380          & 0.5449          & 0.5580          & 0.5952          \\
			& SAPLMA                                       & 0.8170          & 0.8280          & 0.8200          & 0.8315          & 0.8150          & 0.8200          & 0.7799          & 0.7884          \\
			& LLM-Check                                    & 0.6316          & 0.6006          & 0.5552          & 0.5470          & 0.5726          & 0.5773          & 0.5292          & 0.5398          \\
			& ICR-Probe                                    & 0.7937          & 0.7740          & 0.7684          & 0.7560          & 0.7595          & 0.7215          & 0.8003          & 0.7730          \\
			& HaluGNN                                      & 0.8392          & 0.7706          & 0.9050          & 0.8245          & 0.9217          & \textbf{0.8402} & 0.9065          & 0.8058          \\
			& \textbf{CausalGaze}                          & \textbf{0.8680±0.0311} & \textbf{0.8071±0.0102} & \textbf{0.9106±0.0110} & \textbf{0.8280±0.0540} & \textbf{0.9328±0.0021} & 0.8283±0.0052          & \textbf{0.9220±0.0231} & \textbf{0.8360±0.0320} \\ \midrule
			\multirow{11}{*}{Mistral-7B} & SelfCheckGPT                                 & 0.5771          & 0.5345          & 0.6340          & 0.6145          & 0.5593          & 0.5260          & 0.6729          & 0.6357          \\
			& EigenScore                                   & 0.6012          & 0.5860          & 0.6573          & 0.6270          & 0.6474          & 0.6346          & 0.5950          & 0.5659          \\
			& Perplexity                                   & 0.5538          & 0.5421          & 0.5740          & 0.5532          & 0.5470          & 0.5583          & 0.5525          & 0.5244          \\
			& LN-Entropy                                   & 0.5763          & 0.5720          & 0.5834          & 0.5682          & 0.6036          & 0.5832          & 0.7156          & 0.6580          \\
			& Semantic Entropy                             & 0.6557          & 0.6136          & 0.6063          & 0.5770          & 0.6945          & 0.6640          & 0.6685          & 0.7065          \\
			& P(True)                                      & 0.5260          & 0.5125          & 0.5680          & 0.5560          & 0.5673          & 0.5531          & 0.5739          & 0.5643          \\
			& SAPLMA                                       & 0.8112          & 0.7883          & 0.8290          & 0.7984          & 0.7884          & 0.7790          & 0.7880          & 0.7729          \\
			& LLM-Check                                    & 0.6732          & 0.6540          & 0.5417          & 0.5411          & 0.5748          & 0.5547          & 0.5373          & 0.5870          \\
			& ICR-Probe                                    & 0.7993          & \textbf{0.8000} & 0.7325          & 0.7258          & 0.7793          & {0.7826} & 0.8047          & 0.8120          \\
			& HaluGNN                                      & \textbf{0.8647}          & 0.7747          & 0.8735          & 0.8000          & 0.8343          & 0.7273          & 0.8647          & 0.8030          \\
			& \textbf{CausalGaze}                          & {0.8851±0.0092} & 0.7971±0.0075          & \textbf{0.8880±0.0621} & \textbf{0.8113±0.0210} & \textbf{0.8550±0.0023} & \textbf{0.7860±0.0041}          & \textbf{0.9127±0.0382} & \textbf{0.8292±0.0273} \\ 
			\bottomrule
	\end{tabular}}
	\caption{Main results of AUROC and F1-Score compared with different competitive baselines over diverse LLMs on TruthfulQA, TriviaQA, SciQ and HaluEval datasets. The best results are highlighted in bold.}
	\label{tab1}
\end{table*}

\subsection{Implementation Details}
\label{subsec4.5}
The CausalGaze model employs a multi-stage GNN architecture optimized for high-dimensional sequential feature processing. Specifically, the high-dimensional node features are first projected down to a 128-dimensional hidden space via a dedicated projection Layer to reduce the computational overhead and parameter counts. Prior to message passing, we utilize a causal refinement layer to obtain the dynamically refined adjacency matrix. The 128-dimensional features are processed by two stacked multi-head GAT layers (with 4 heads) to adaptively learn causal relationships. A hybrid pooling layer concatenating the global max pooling and mean pooling forms a 256-dimensional embedding, which is passed through an MLP classifier for binary prediction. The binary cross-entropy loss with a regularization term is applied to the sigmoid of the model outputs.

\section{Experimental Results and Analysis}
\subsection{Main Results}
\label{subsec5.1}

The main experimental results are demonstrated in Table~\ref{tab1}, showing that the CausalGaze has competitive performance against most baseline methods on most datasets and diverse LLMs. 

Specifically, the experimental results indicate that classification-based approaches substantially outperform other baseline methods. This superiority is primarily because baseline methods rely exclusively on the LLM’s intrinsic mechanisms, whereas classification-based approaches leverage an additionally trained detection probe. While existing classification-based methods, such as SAPLMA, LLM-Check, and ICR Probe, utilize either hidden states or attention maps in isolation and yield sound detection performance, HaluGNN effectively achieves 5-7\% improvement by coupling both hidden states and attention maps into a graph structure. Nevertheless, HaluGNN directly utilize attention maps as graph edges, inherently introducing a substantial amount of noise and irrelevant connections, as illustrated in Figure~\ref{example} and \ref{gradient}. This limits the classifier’s capacity to learn beyond superficial correlational relationships. In contrast, our proposed CausalGaze framework not only mitigates this issue by substantially reducing noise and spurious connections but also fortifies the underlying causal relationship between the generated content and the factual or hallucinated labels. Compared with HaluGNN, our approach achieves over 3.3\% improvement in AUROC on the TruthfulQA dataset. More details are shown in Appendix~\ref{appendixBadd}.

\subsection{Generalization Analysis}
\label{subsec5.2}

\begin{figure}[t]
	\centering 
	\includegraphics[width=\columnwidth]{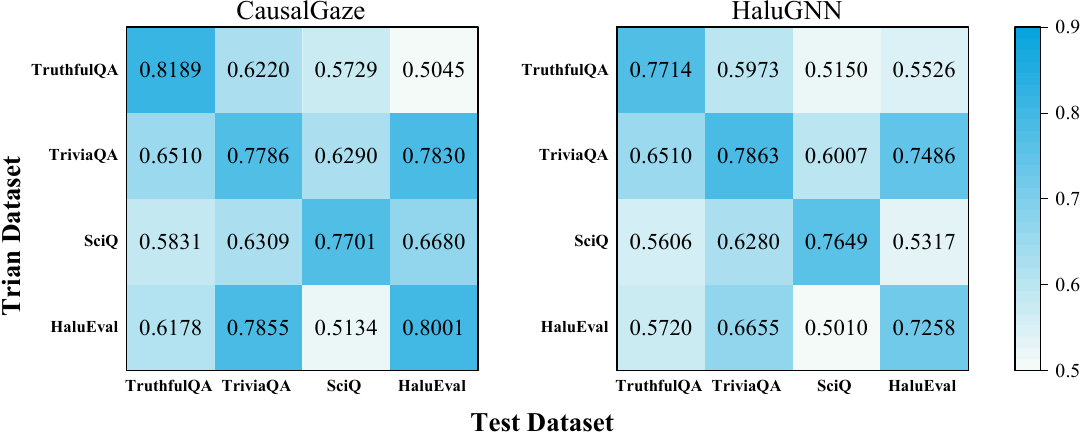}
	\caption{Cross-dataset generalization analysis for CausalGaze and HaluGNN. Each subplot displays the F1-Score when the model is trained on the row dataset and tested on the column dataset, with values annotated in each cell.}
	\label{cross}
\end{figure}

To assess the cross-dataset generalization capability, we train each model on one dataset and evaluate its performance on all other datasets. The results are visualized as a heatmap in Figure~\ref{cross}, where Each cell in the heatmap reports the F1-Score, sharing the same color scale (ranging from 0.5 to 0.9) for comparison.

On unseen target datasets, our proposed CausalGaze consistently and significantly outperforms HaluGNN. Furthermore, the CausalGaze evaluation on target-domain improves the average F1-Score by 3.6\% compared to HaluGNN, which demonstrates the superior robustness of CausalGaze under domain shift. This notable advantage stems from CausalGaze capturing factual and hallucinated patterns from the causal perspective. 

% Please add the following required packages to your document preamble:
% \usepackage{multirow}
\begin{table*}[t]
	\centering
	
	\resizebox{\textwidth}{!}{
		\begin{tabular}{cccccccccc}
			\toprule
			\multirow{2}{*}{\textbf{Projection   Layer}} & \multirow{2}{*}{\textbf{Dimension}} & \multicolumn{2}{c}{\textbf{TruthfulQA}} & \multicolumn{2}{c}{\textbf{TriviaQA}} & \multicolumn{2}{c}{\textbf{SciQ}} & \multicolumn{2}{c}{\textbf{Halueval}} \\ \cline{3-10} 
			&                                     & AUROC              & F1-Score           & AUROC             & F1-Score          & AUROC           & F1-Score        & AUROC             & F1-Score          \\ \midrule
			w/o                                          & 4096                                   & 0.6069             & 0.6469             & 0.5898            & 0.5318            & 0.5889          & 0.6559          & 0.5530            & 0.6676            \\ \midrule
			\multirow{4}{*}{with}                        & 64                                  & 0.8712             & 0.7818             & 0.8719            & 0.7809            & 0.8301          & \textbf{0.7773}          & 0.8989            & 0.6322            \\
			& 128                                 & {0.8805}    & \textbf{0.8189}    & 0.8614            & 0.7786            & \textbf{0.8390} & {0.7701} & \textbf{0.9280}   & \textbf{0.8001}   \\
			& 256                                 & 0.8413             & 0.7822             & 0.8840            & \textbf{0.8011}   & 0.8324          & 0.7462          & 0.8837            & 0.7093            \\
			& 512                                 & \textbf{0.8807}             & 0.8018             & \textbf{0.8849}   & 0.7956            & 0.8267          & 0.7333          & 0.8168            & 0.7698            \\ \bottomrule
	\end{tabular}}
	\caption{The contribution of the Projection Layer and its dimentions on Llama2-7B across different datasets. The best results are highlighted in bold.}
	\label{tab2}
\end{table*}

\begin{figure}[t]
	\centering 
	\includegraphics[width=\columnwidth]{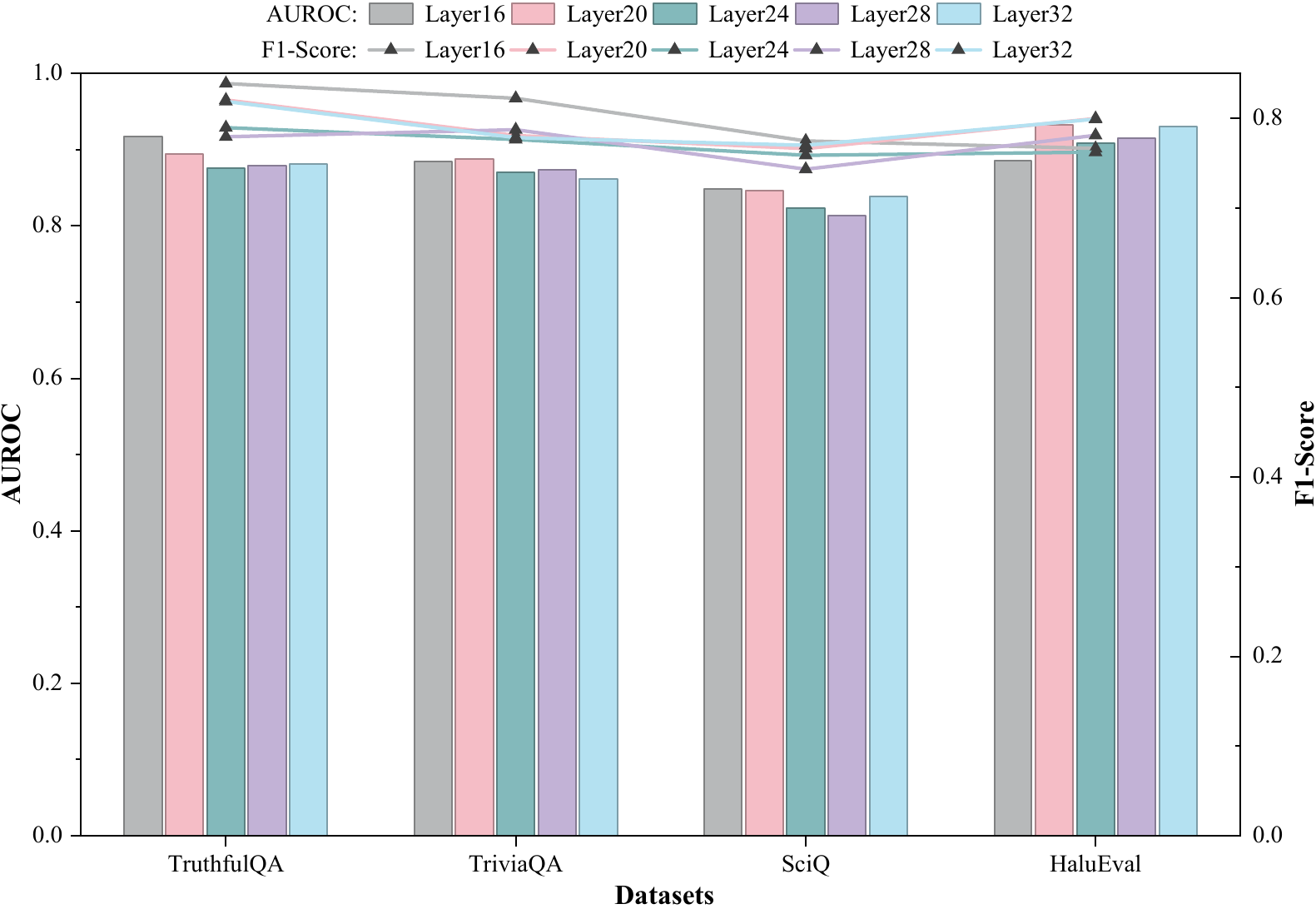}
	\caption{The impact of causal graphs from different layers of Llama2-7B on the detection performance.}
	\label{layer}
\end{figure}

\begin{table}[t]
	\centering
	\resizebox{\columnwidth}{!}{
		\begin{tabular}{llcccc}
			\hline
			\multicolumn{1}{c}{\textbf{Model}}             & \textbf{Variant} & \textbf{TruthfulQA} & \textbf{TriviaQA} & \textbf{SciQ}   & \textbf{HaluEval} \\ \hline
			\multicolumn{1}{c}{\multirow{4}{*}{Llama2-7B}} & CausalGaze       & \textbf{0.8189}     & \textbf{0.7786}   & \textbf{0.7701} & \textbf{0.8001}   \\
			\multicolumn{1}{c}{}                           & w/o Gradient     & 0.7720              & 0.7763            & 0.7549          & 0.6316            \\
			\multicolumn{1}{c}{}                           & Random Gradient  & 0.7811              & 0.7656            & 0.7319          & 0.6630            \\
			\multicolumn{1}{c}{}                           & MLP($A$)         & 0.7793              & 0.7674            & 0.7540          & 0.7241            \\ \hline
			\multirow{4}{*}{Qwen2-7B}                      & CausalGaze       & \textbf{0.8071}     & \textbf{0.8280}   & \textbf{0.8283} & \textbf{0.8360}   \\
			& w/o Gradient     & 0.7660              & 0.8240            & 0.8400          & 0.8033            \\
			& Random Gradient  & 0.7672              & 0.8252            & 0.8233          & 0.6667            \\
			& MLP($A$)         & 0.7857              & 0.8246            & 0.8520          & 0.6947            \\ \hline
			\multirow{4}{*}{Mistral-7B}                    & CausalGaze       & \textbf{0.7971}     & \textbf{0.8113}   & \textbf{0.7860} & \textbf{0.8292}   \\
			& w/o Gradient     & 0.7746              & 0.7992            & 0.7283          & 0.8020            \\
			& Random Gradient  & 0.7636              & 0.7927            & 0.7787          & 0.7623            \\
			& MLP($A$)        & 0.7811              & 0.7908            & 0.7407          & 0.7855            \\ \hline
	 \end{tabular}}
	\caption{The performance comparison of CausalGaze with w/o Gradient, Random Gradient and A learnable gating mechanism based solely on the attention maps ($A$) (i.e., MLP($A$)).}
	\label{add1}
\end{table}

\begin{figure*}[t]
	\centering
	% 第一张图
	\begin{minipage}[b]{0.47\textwidth}
		\centering
		\includegraphics[width=\textwidth]{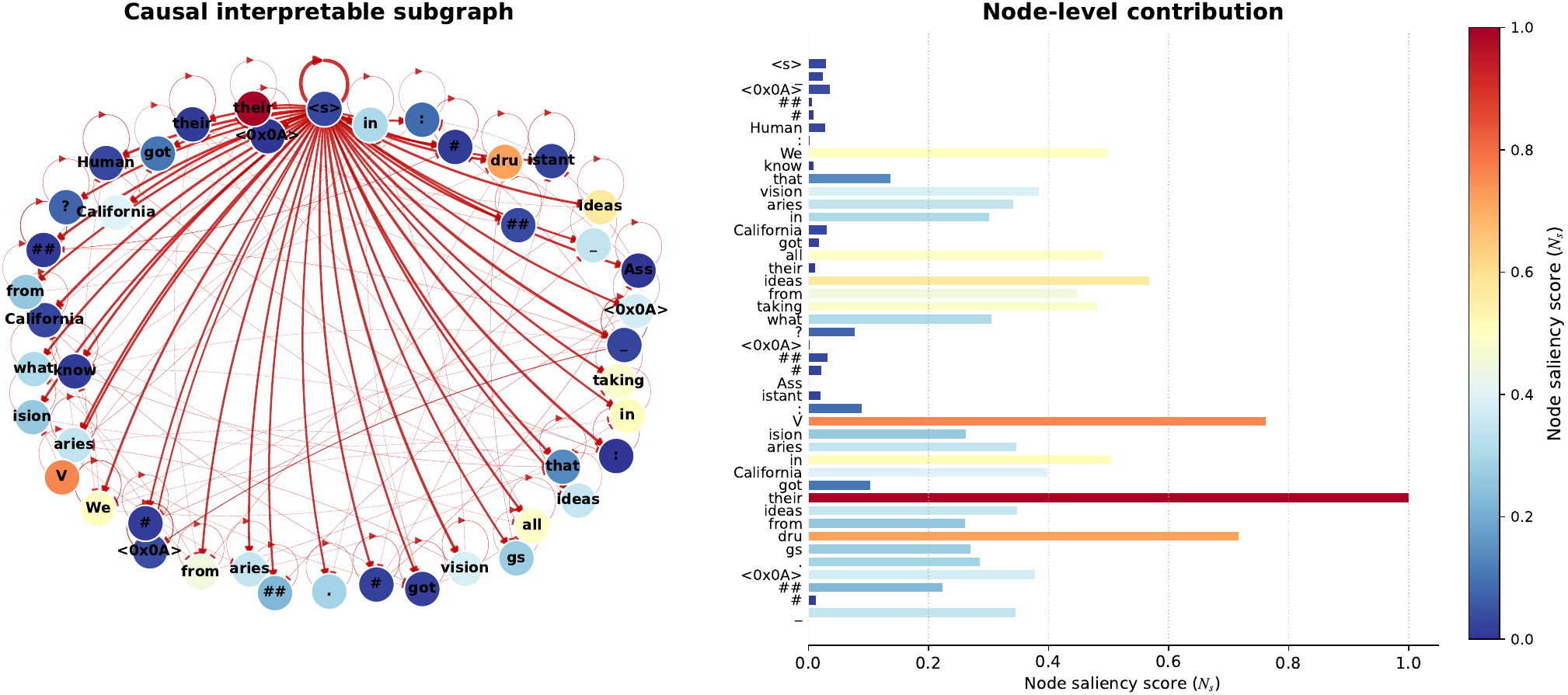}
		\caption*{(a) The causal nodes and edges for the hallucination.} % 如果不需要子标题，可以删掉
	\end{minipage}
	\label{a}
	\hfill% 在两图之间插入弹性间距
	% 第二张图
	\begin{minipage}[b]{0.47\textwidth}
		\centering
		\includegraphics[width=\textwidth]{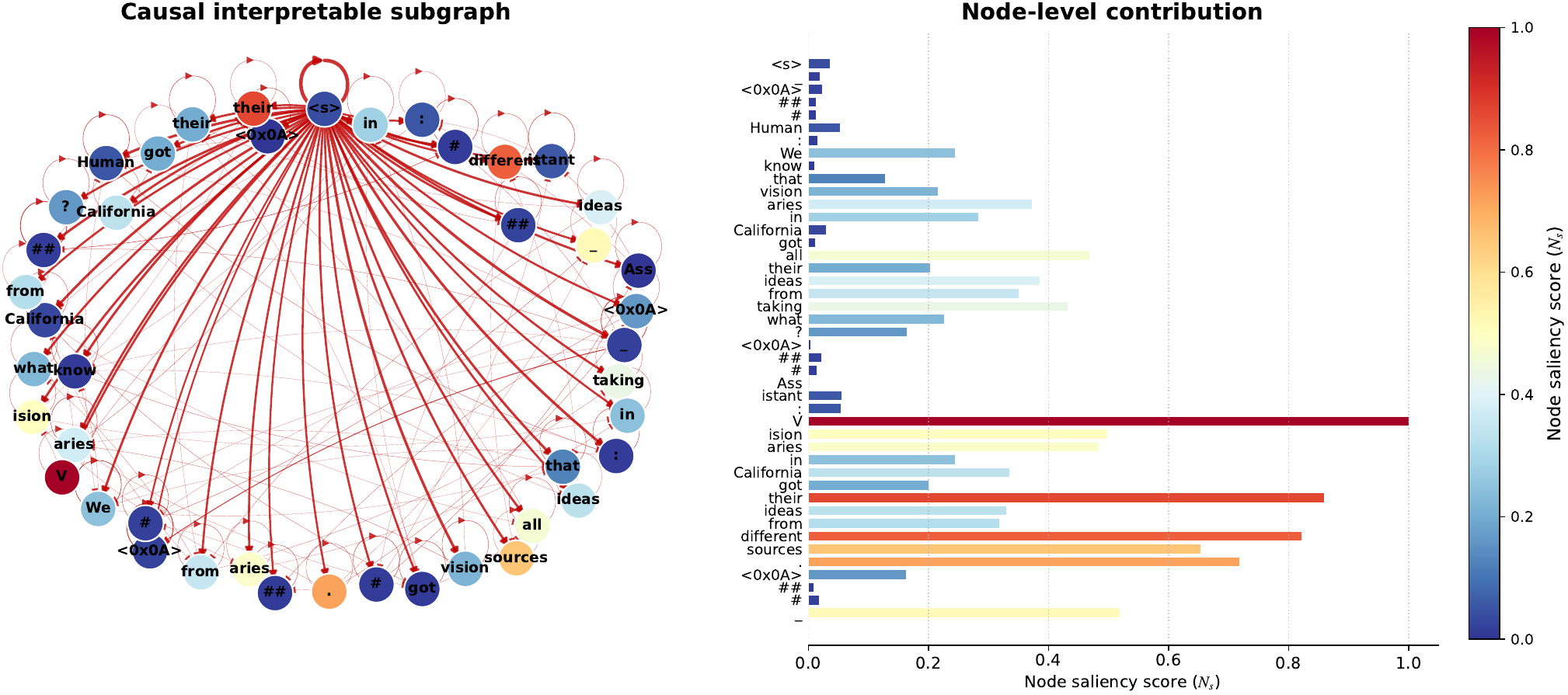}
		\caption*{(b) The causal nodes and edges for the fact.}
	\end{minipage}
	\label{b}
	\caption{The token-level and fine-grained interpretability analysis of the detection result for the hallucinated and factual responses of the same question in Figure~\ref{example}.}
	\label{Interpretability}
\end{figure*}

\subsection{Ablation Experiments and Interpretability}
\label{subsec5.3}
We conduct ablation experiments on Llama-2-7B \citep{add1} model, mainly studying three aspects: (1) The contribution of projection layer and its dimension to the detection performance; (2) The effect of different layers on the detection performance; (3) The effect of gradient-guided information. More details are shown in Appendix~\ref{appendixB}.

\textbf{Contribution of the Projection Layer and its Dimension.} While the projection layer is introduced to the CausalGaze architecture primarily to decrease the high-dimensional node features, it is necessary to experimentally investigate the impact of both the dimensionality reduction operation itself and the choice of the compressed dimension on the detection performance. Given that the padded hidden state dimension of Llama2-7B model is 4096, we conduct an ablation study to test four representative dimensions: 64, 128, 256, and 512. The experimental results, summarized in Table~\ref{tab2}, confirm that the dimensionality reduction strategy is absolutely effective for improving detection performance. Furthermore, the dimension of 128 strikes an optimal balance, achieving the competitive detection performance while maintaining a relatively minimal model parameter count.

\textbf{Layer Number.} Prior literature suggest that LLMs gradually capture and understand context, and ensure token fluency primarily within the initial layers. Subsequently, the middle and later layers are responsible for knowledge integration, next-token generation, and implicitly contain more causal information related to factual consistency. To investigate the optimal source for the proposed CausalGaze, we conduct an ablation study using causal graphs extracted from five layers of the Llama2-7B model: 16, 20, 24, 28, and 32. The results, as illustrated in Figure~\ref{layer}, indicate that features derived from Layer 20 yield the best performance. This finding aligns with existing research, where critical semantic and factual consistency features relevant to the generated content are primarily found in the middle-to-later layers of the LLMs \citep{bib24,bib12,bib14}.

\textbf{Contribution of the Gradient-guided Information.} The gradient-guided sensitivity $S$ is actually the cornerstone of CausalGaze. To isolate and verify its specific contribution, we conduct an additional ablation study to compare the full model with three variants that bypass the gradient-guided information: (1) w/o Gradients ($S$), where we replace the gradient-guided sensitivity matrix with a uniform matrix in Equation \ref{eq8}. It means the gating mechanism relies solely on the raw attention weights without any causal sensitivity guidance; (2) Random Gradients, where we replace with a randomly initialized matrix to test if any additional information or random perturbation could provide a similar regularization effect; (3) A learnable gating mechanism based solely on the attention maps ($A$) (i.e., MLP($A$)). The comparison of F1-Score on all datasets and models is shown in Table \ref{add1}. The significant performance drop (about 2-4\%) when removing the gradients or setting random gradients confirms that the learned gradients $S$ is indeed the key driver of our model's competitive performance. While the MLP($A$) improves slightly over the w/o gradients, it still significantly underperforms CausalGaze. This proves that raw attention $A$ may only capture statistical correlations, which are often the source of spurious correlations in LLM hallucinations, whereas gradients $S$ identify the causal bottlenecks. Also, the gradient-guided gradients $\nabla_{{A}}\mathcal{L}$ reflect how a specific attention edge contributes to the final results, and an edge can have a high attention weight (statistical significance) but low causal sensitivity (not the cause of hallucination or fact). By integrating the gradient-guided information, CausalGaze specifically detects hallucination with interpretability, where a simple MLP($A$) or random gradients cannot make it.

\textbf{Token-level Interpretability.} To address the inherent lack of interpretability in existing classification-based hallucination detection methods, CausalGaze enables fine-grained and token-level analysis grounded in causal intervention. We perform node and edge importance analysis on the Llama2-7B model using a representative example from TruthfulQA (seen in Figure~\ref{example}), and the visualization is depicted in Figure~\ref{Interpretability}. It can be evidently observed that the causal interpretable graphs successfully mitigate noise and prune spurious connections. Specifically, the importance scores of nodes representing key entities (such as `Visionaries', `ideas' and `drugs' in Figure~\ref{Interpretability}(a) or `sources' in Figure~\ref{Interpretability}(b)) are significantly higher than those assigned to non-informative tokens (such as `\#\#', `<s>', and `<0x0A>'). Crucially, the edge from token `ideas' to `drugs' is severed in the hallucinated answer, whereas the edge from token `ideas' to `sources' remains intact in the fact answer, which is consistent with our theoretical expectations. We attribute this disparity to the fact that while such tokens may exhibit high co-occurrence in the pre-training corpus, they lack genuine causal dependencies; consequently, the models may prioritize linguistic fluency over factual accuracy, leading to the risky guesses \citep{add8}. By identifying the causal structures critical to detection, CausalGaze effectively bridges the gap in model interpretability.

\section{Conclusion}
\label{sec5}
In this paper, we propose CausalGaze, the first work to introduce active causal intervention to address the passive attribution in hallucination detection tasks. CausalGaze effectively suppresses the noise and spurious correlations inherent in raw representations, enabling a precise probing of the causal relationships between internal signals and model outputs. Comprehensive experimental results across multiple benchmarks demonstrate that CausalGaze not only achieves competitive performance over existing methods but also provides a robust, interpretable foundation for hallucination detection.

%% Use \subsection commands to start a subsection.

\section*{Limitations}

While the proposed CausalGaze shows significantly effective performance and bridges the gap in model interpretability, it is subject to certain limitations. Firstly, the method necessitates access to the LLMs' latent space, which inherently restricts its application in proprietary and black-box LLMs. Secondly, our approach only models the local causal graph from a single layer, which can not capture the globally comprehensive information embedded across all layers of LLMs. And the graph computation process is resource-intensive, which limits its practical use in the detection of long-context texts. Building upon the insights, we anticipate that future research will explore more efficient and alternative signal representation paradigms, such as aggregating features from multiple layers or using summary vectors, to effectively expand the applicability and scalability of causal intervention methods for hallucination detection.

\section*{Ethics and Broader Impact}

We sampled a portion of the data from existing datasets for our experiments, which may affect the accuracy of some of our conclusions.

% Bibliography entries for the entire Anthology, followed by custom entries
%\bibliography{custom,anthology-overleaf-1,anthology-overleaf-2}

% Custom bibliography entries only
\bibliography{custom}

@inproceedings{Lookback,
	title = {Lookback {Lens}: {Detecting} and {Mitigating} {Contextual} {Hallucinations} in {Large} {Language} {Models} {Using} {Only} {Attention} {Maps}},
	doi = {10.18653/v1/2024.emnlp-main.84},
	publisher = {Proceedings of the 2024 Conference on Empirical Methods in Natural Language Processing},
	author = {Chuang, Yung-Sung and Qiu, Linlu and Hsieh, Cheng-Yu and Krishna, Ranjay and Kim, Yoon and Glass, James R.},
	year = {2024},
	pages = {1419--1436},
}

@inproceedings{Factoscope,
	title = {{LLM} {Factoscope}: {Uncovering} {LLMs}' {Factual} {Discernment} through {Inner} {States} {Analysis}},
	doi = {10.18653/v1/2024.findings-acl.608},
	author = {He, Jinwen and Gong, Yujia and Chen, Kai and Lin, Zijin and Wei, Chengan and Zhao, Yue},
	year = {2024},
	booktitle = {Findings of the 62nd Annual Meeting of the Association for Computational Linguistics, ACL 2024},
}

@inproceedings{TAD,
	title = {Unconditional {Truthfulness}: {Learning} {Unconditional} {Uncertainty} of {Large} {Language} {Models}},
	author = {Vazhentsev, Artem and Fadeeva, Ekaterina and Xing, Rui and Kuzmin, Gleb and Lazichny, Ivan and Panchenko, Alexander and Nakov, Preslav and Baldwin, Timothy and Panov, Maxim and Shelmanov, Artem},
	year = {2025},
	doi = {10.18653/v1/2025.emnlp-main.1807},
	booktitle = {Proceedings of the 2025 Conference on Empirical Methods in Natural Language Processing, EMNLP 2025},

}

@inproceedings{add4,
    title = "Leveraging Graph Structures to Detect Hallucinations in Large Language Models",
    author = "Nonkes, Noa and Agaronian, Sergei and Kanoulas, Evangelos and Petcu, Roxana",
    booktitle = {Findings of the {Association} for {Computational} {Linguistics} {ACL} 2024},
    year = "2024",
    doi = {10.18653/v1/2024.textgraphs-1.7 },
    pages = "93--104",
}

@inproceedings{bib6,
	title = {{SelfCheckGPT}: {Zero}-{Resource} {Black}-{Box} {Hallucination} {Detection} for {Generative} {Large} {Language} {Models}},
	author = {Manakul, Potsawee and Liusie, Adian and Gales, Mark J. F.},
	year = {2023},
	doi = {10.18653/v1/2023.emnlp-main.557 },
	booktitle= {Proceedings of the 2023 Conference on Empirical Methods in Natural Language Processing}, 
	pages = "9004--9017" 
}

@inproceedings{bib8,
	title = {{INSIDE}: {LLMs}' {Internal} {States} {Retain} the {Power} of {Hallucination} {Detection}},
	booktitle = {12th International Conference on Learning Representations, ICLR 2024},
	author = {Chen, Chao and Liu, Kai and Chen, Ze and Gu, Yi and Wu, Yue and Tao, Mingyuan and Fu, Zhihang and Ye, Jieping},
	year = {2024},
    url = {https://openreview.net/forum?id=Zj12nzlQbz},
}

@inproceedings{bib24,
	title = {The {Internal} {State} of an {LLM} {Knows} {When} {It}'s {Lying}},
	doi = {10.18653/v1/2023.findings-emnlp.68 },
	author = {Azaria, Amos and Mitchell, Tom},
	year = {2023},
	booktitle= {Findings of the Association for Computational Linguistics: EMNLP 2023}, 
	pages = "967--976" 

}

@inproceedings{bib10,
	title = {Unsupervised {Real}-{Time} {Hallucination} {Detection} based on the {Internal} {States} of {Large} {Language} {Models}},
	doi = {10.18653/v1/2024.findings-acl.854 },
	booktitle = "Findings of the Association for Computational Linguistics: ACL 2024",
	author = {Su, Weihang and Wang, Changyue and Ai, Qingyao and HU, Yiran and Wu, Zhijing and Zhou, Yujia and Liu, Yiqun},
	year = {2024},
	pages = "14379--14391"
}

@inproceedings{bib11,
	author = {Xuefeng Du and Chaowei Xiao and Yixuan Li},
	title ={{HaloScope}: {Harnessing} {Unlabeled} {LLM} {Generations} for {Hallucination} {Detection}},
	doi = { 10.18653/v1/d19-1410 },
	booktitle={Proceedings of the 38th International Conference on Neural Information Processing Systems, NeurIPS 2024},
	year={2024},
}

@inproceedings{bib12,
	title = {LLM-Check: Investigating Detection of Hallucinations in Large Language Models},
	booktitle = {38th Conference on Neural Information Processing Systems, NeurIPS 2024},
	author = {Sriramanan, Gaurang and  Bharti, Siddhant and  Sadasivan, Vinu Sankar and  Saha, Shoumik and  Kattakinda, Priyatham and  Feizi},
    doi = {10.52202/079017-1077},
    year={2024},
}

@inproceedings{bib13,
	title={ICR Probe: Tracking Hidden State Dynamics for Reliable Hallucination Detection in LLMs},
	author={Zhenliang Zhang and Xinyu Hu and Huixuan Zhang and Junzhe Zhang and Xiaojun Wan},
	booktitle={63rd Annual Meeting of the Association-for-Computational-Linguistics, ACL 2025.},
    doi ={10.18653/v1/2025.acl-long.880},
    year={2025},
	pages = "17986--18002"
}

@inproceedings{bib15,
    title={(Im)possibility of Automated Hallucination Detection in Large Language Models},
    author={Karbasi, Amin and  Montasser, Omar and  Sous, John and  Velegkas, Grigoris},
    booktitle={CONFERENCE ON LANGUAGE MODELING, COLM 2025},
    year={2025},
}

@inproceedings{bib25,
    title={GNNExplainer: Generating Explanations for Graph Neural Networks},
    author={Ying, Rex and Bourgeois, Dylan and You, Jiaxuan and Zitnik, Marinka and Leskovec, Jure},
    booktitle={33rd Annual Conference on Neural Information Processing Systems, NeurIPS 2019},
    year={2019},
    url={https://www.engineeringvillage.com/app/doc/?docid=cpx_M441df12c174920cac9eM6ff410178163190}
}

@inproceedings{bib26,
    title={Parameterized explainer for graph neural network},
    author={Dongsheng Luo and Wei Cheng and Dongkuan Xu and Wenchao Yu and Bo Zong and Haifeng Chen and Xiang Zhang},
    booktitle={34th Conference on Neural Information Processing Systems, NeurIPS 2020},
    year={2020},
    url={https://dl.acm.org/doi/10.5555/3495724.3497370}
}

@conference{add5,
	title = {Uncertainty estimation in autoregressive structured prediction},
	booktitle = {9th International Conference on Learning Representations, ICLR 2021},
	author = {Malinin, A. and Gales, M.},
	year = {2021},
}

@conference{add6,
	title = {Out-of-{Distribution} {Detection} and {Selective} {Generation} for {Conditional} {Language} {Models}},
	booktitle = {11th International Conference on Learning Representations, ICLR 2023},
	author = {Ren, Jie and Luo, Jiaming and Zhao, Yao and Krishna, Kundan and Saleh, Mohammad and Lakshminarayanan, Balaji and Liu, Peter J.},
	year = {2023},
    
}

@inproceedings{bib17,
    title={FRED: Financial Retrieval-Enhanced Detection and Editing of Hallucinations in Language Models},
    author={Likun Tan and Kuan-Wei Huang and Kevin Wu},
    booktitle={Proceedings of the 42nd International Conference on Machine Learning, ICML 2025},
    url={https://arxiv.org/abs/2507.20930},
    year={2025},
}

@inproceedings{bib18,
    title={DAHL: Domain-specific Automated Hallucination Evaluation of Long-Form Text through a Benchmark Dataset in Biomedicine},
    author={Seo, Jean and Lim, Jongwon and Jang, Dongjun and Shin, Hyopil},
    booktitle={The 2024 Conference on Empirical Methods in Natural Language Processing, EMNLP 2024},
    url={https://arxiv.org/abs/2411.09255},
    year={2024},
}

@inproceedings{bib19,
    title={HD-NDEs: Neural Differential Equations for Hallucination Detection in LLMs},
    author={Qing Li and Jiahui Geng and Zongxiong Chen and Derui Zhu and Yuxia Wang and Congbo Ma and Chenyang Lyu and Fakhri Karray},
    booktitle={Proceedings of the 63rd Annual Meeting of the Association for Computational Linguistics (Volume 1: Long Papers)},
    doi = "10.18653/v1/2025.acl-long.309",
    pages = "6173--6186",
    year={2025},
}

@inproceedings{bib23,
    title = "Hallucination Detection in {LLM}s Using Spectral Features of Attention Maps",
    author = "Binkowski, Jakub and Janiak, Denis and Sawczyn, Albert and Gabrys, Bogdan and Kajdanowicz, Tomasz Jan",
    booktitle = "Proceedings of the 2025 Conference on Empirical Methods in Natural Language Processing",
    year = "2025",
    publisher = acl,
    doi = "10.18653/v1/2025.emnlp-main.1239",
    pages = "24365--24396",
}

@inproceedings{bib21,
	title = {Out-of-{Distribution} {Detection} and {Selective} {Generation} for {Conditional} {Language} {Models}},
	booktitle = {11th International Conference on Learning Representations, ICLR 2023},
	author = {Ren, Jie and Luo, Jiaming and Zhao, Yao and Krishna, Kundan and Saleh, Mohammad and Lakshminarayanan, Balaji and Liu, Peter J.},
	year = {2023},
    
}

@inproceedings{bib22,
	title={SEMANTIC UNCERTAINTY: LINGUISTIC INVARIANCES FOR UNCERTAINTY ESTIMATION IN NATURAL LANGUAGE GENERATION},
	author={Kuhn, Lorenz and Gal, Yarin and Farquhar, Sebastian},
	booktitle={11th International Conference on Learning Representations, ICLR 2023},
	year={2023},
}

@inproceedings{bib28,
	title={TruthfulQA: Measuring How Models Mimic Human Falsehoods},
	author={Lin, Stephanie and Hilton, Jacob and Evans, Owain},
	booktitle={Proceedings of the 60th Annual Meeting of the Association for Computational Linguistics (Volume 1: Long Papers)},
	year={2022},
	doi={10.18653/v1/2022.acl-long.229},
	pages = "3214--3252"
}

@inproceedings{bib29,
	title={TriviaQA: A Large Scale Distantly Supervised Challenge Dataset for Reading Comprehension},
	author={Mandar Joshi and Eunsol Choi and Daniel S. Weld},
	booktitle={Proceedings of the 55th Annual Meeting of the Association for Computational Linguistics (Volume 1: Long Papers)},
	doi={10.18653/v1/p17-1147 },
	year={2017},
	pages = "1601--1611"
}

@article{bib30,
	title={Crowdsourcing Multiple Choice Science Questions},
	author={Johannes Welbl and Nelson F. Liu and Matt Gardner},
	journal={Statistics},
	doi={10.18653/v1/w17-4413 },
	year={2017},
}

@inproceedings{bib31,
	title = "{H}alu{E}val: A Large-Scale Hallucination Evaluation Benchmark for Large Language Models",
	author = "Li, Junyi and Cheng, Xiaoxue and Zhao, Xin and Nie, Jian-Yun and Wen, Ji-Rong",
	booktitle = "Proceedings of the 2023 Conference on Empirical Methods in Natural Language Processing",
	year = "2023",
	publisher = acl,
	doi = "10.18653/v1/2023.emnlp-main.397",
	pages = "6449--6464"
}

@preprints{add7,
	title= {Language Models (Mostly) Know What They Know},
	author={Saurav Kadavath and Tom Conerly and Amanda Askell and Tom Henighan and Dawn Drain and Ethan Perez and Nicholas Schiefer and Zac Hatfield-Dodds and Nova DasSarma and Eli Tran-Johnson and Scott Johnston and Sheer El-Showk and Andy Jones and Nelson Elhage and Tristan Hume and Anna Chen and Yuntao Bai and Sam Bowman and Stanislav Fort and Deep Ganguli and Danny Hernandez and Josh Jacobson and Jackson Kernion and Shauna Kravec and Liane Lovitt and Kamal Ndousse and Catherine Olsson and Sam Ringer and Dario Amodei and Tom Brown and Jack Clark and Nicholas Joseph and Ben Mann and Sam McCandlish and Chris Olah and Jared Kaplan},
	howpublished  = {arXiv preprint arXiv:2207.05221},
	year={2022},
	url={https://arxiv.org/abs/2207.05221}
}

@preprints{bib9, 
	title = {Siren's {Song} in the {AI} {Ocean}: {A} {Survey} on {Hallucination} in {Large} {Language} {Models}},
	howpublished  = {arXiv preprint arXiv:2309.01219},
	author = {Zhang, Yue and Li, Yafu and Cui, Leyang and Cai, Deng and Liu, Lemao and Fu, Tingchen and Huang, Xinting and Zhao, Enbo and Zhang, Yu and Chen, Yulong and Wang, Longyue and Luu, Anh Tuan and Bi, Wei and Shi, Freda and Shi, Shuming},
	year = {2023},
    url = {https://arxiv.org/abs/2502.17598},
}

@preprints{bib5,
	title = {Addressing {Hallucinations} in {Language} {Models} with {Knowledge} {Graph} {Embeddings} as an {Additional} {Modality}},
	author = {Chekalina, Viktoriia and Razzhigaev, Anton and Goncharova, Elizaveta and Kuznetsov, Andrey},
	howpublished  = {arXiv preprint arXiv:2411.11531},
	year = {2025},
    url = {https://arxiv.org/abs/2411.11531},
}

@preprints{add1,
	title={Llama 2: Open Foundation and Fine-Tuned Chat Models},
	howpublished  = {arXiv preprint arXiv:2307.09288},
	author={Touvron, Hugo and  Martin, Louis and  Stone, Kevin and  Albert, Peter and  Almahairi, Amjad and  Babaei, Yasmine and  Bashlykov, Nikolay and  Batra, Soumya and  Bhargava, Prajjwal and  Bhosale, Shruti and  Bikel, Dan and  Blecher, Lukas and  Ferrer, Cristian Canton and  Chen, Moya and  Cucurull, Guillem and  Esiobu, David and  Fernandes, Jude and  Fu, Jeremy and  Fu, Wenyin and  Fuller, Brian and  Gao, Cynthia and  Goswami, Vedanuj and  Goyal, Naman and  Hartshorn, Anthony and  Hosseini, Saghar and  Hou, Rui and  Inan, Hakan and  Kardas, Marcin and  Kerkez, Viktor and  Khabsa, Madian and  Kloumann, Isabel and  Korenev, Artem and  Koura, Punit Singh and  Lachaux, Marie-Anne and  Lavril, Thibaut and  Lee, Jenya and  Liskovich, Diana and  Lu, Yinghai and  Mao, Yuning and  Martinet, Xavier and  Mihaylov, Todor and  Mishra, Pushkar and  Molybog, Igor and  Nie, Yixin and  Poulton, Andrew and  Reizenstein, Jeremy and  Rungta, Rashi and  Saladi, Kalyan and  Schelten, Alan and  Silva, Ruan and  Smith, Eric Michael and  Subramanian, Ranjan and  Tan, Xiaoqing Ellen and  Tang, Binh and  Taylor, Ross and  Williams, Adina and  Kuan, Jian Xiang and  Xu, Puxin and  Yan, Zheng and  Zarov, Iliyan and  Zhang, Yuchen and  Fan, Angela and  Kambadur, Melanie and  Narang, Sharan and  Rodriguez, Aurelien and  Stojnic, Robert and  Edunov, Sergey and  Scialom, Thomas},
	year={2023},
    url={https://arxiv.org/abs/2307.09288}
}

@preprints{add2,
	title={Qwen2 Technical Report},
	author={An Yang and Baosong Yang and Binyuan Hui and Bo Zheng and Bowen Yu and Chang Zhou and Chengpeng Li and Chengyuan Li and Dayiheng Liu and Fei Huang and Guanting Dong and Haoran Wei and Huan Lin and Jialong Tang and Jialin Wang and Jian Yang and Jianhong Tu and Jianwei Zhang and Jianxin Ma and Jianxin Yang and Jin Xu and Jingren Zhou and Jinze Bai and Jinzheng He and Junyang Lin and Kai Dang and Keming Lu and Keqin Chen and Kexin Yang and Mei Li and Mingfeng Xue and Na Ni and Pei Zhang and Peng Wang and Ru Peng and Rui Men and Ruize Gao and Runji Lin and Shijie Wang and Shuai Bai and Sinan Tan and Tianhang Zhu and Tianhao Li and Tianyu Liu and Wenbin Ge and Xiaodong Deng and Xiaohuan Zhou and Xingzhang Ren and Xinyu Zhang and Xipin Wei and Xuancheng Ren and Xuejing Liu and Yang Fan and Yang Yao and Yichang Zhang and Yu Wan and Yunfei Chu and Yuqiong Liu and Zeyu Cui and Zhenru Zhang and Zhifang Guo and Zhihao Fan},
	howpublished  = {arXiv preprint arXiv:2407.10671},
	year={2024},
    url={https://arxiv.org/abs/2407.10671}
}

@preprints{add3, 
	title = {Mistral 7B},
	howpublished  = {arXiv preprint arXiv:2310.06825},
	author = {Albert Q. Jiang and Alexandre Sablayrolles and Arthur Mensch and et.al},
	year = {2023},
    url={https://arxiv.org/abs/2310.06825}
}

@preprints{add8,
	title={H-Neurons: On the Existence, Impact, and Origin of Hallucination-Associated Neurons in LLMs}, 
	author={Cheng Gao and Huimin Chen and Chaojun Xiao and Zhiyi Chen and Zhiyuan Liu and Maosong Sun},
	year={2025},
	howpublished  = {arXiv preprint arXiv:2512.01797},
	url={https://arxiv.org/abs/2512.01797}, 
}

@article{zhang2026stable,
	title={Stable-RAG: Mitigating Retrieval-Permutation-Induced Hallucinations in Retrieval-Augmented Generation},
	author={Zhang, Qianchi and Zhang, Hainan and Pang, Liang and Zheng, Hongwei and Zheng, Zhiming},
	journal={arXiv preprint arXiv:2601.02993},
	year={2026}
}

@article{bib1,
	title={Hallucination Detection in Foundation Models for Decision-Making: A Flexible Definition and Review of the State of the Art},
	author={Neeloy Chakraborty and Melkior Ornik and Katherine Driggs-Campbell},
	journal={ACM Computing Surveys},
	volume={7},
	pages={1-35},
	year={2025},
    doi = "10.1145/3716846",
}

@article{bib2,
	title = {A {Survey} on {Hallucination} in {Large} {Language} {Models}: {Principles}, {Taxonomy}, {Challenges}, and {Open} {Questions}},
	volume = {43},
	doi = {10.1145/3703155},
	journal = {ACM Transactions on Information Systems},
	author = {Huang, Lei and Yu, Weijiang and Ma, Weitao and Zhong, Weihong and Feng, Zhangyin and Wang, Haotian and Chen, Qianglong and Peng, Weihua and Feng, Xiaocheng and Qin, Bing and Liu, Ting},
	year = {2025},
}

@article{bib3,
	author = {Xiang Shi and Jiawei Liu and Yinpeng Liu and Qikai Cheng and Wei Lu},
	title = {Know where to go: Make LLM a relevant, responsible, and trustworthy searchers},
	journal = {Decision Support Systems},
	volume = {188},
	pages = {114354},
	year = {2025},
	doi = {10.1016/j.dss.2024.114354},
}

@article{bib4,
	title = {HaluCheck: Explainable and verifiable automation for detecting hallucinations in LLM responses},
	journal = {Expert Systems with Applications},
	volume = {272},
	pages = {126712},
	year = {2025},
	doi = {10.1016/j.eswa.2025.126712},
	author = {Sangwoo Heo and Sungwook Son and Hyunwoo Park},
}

@article{bib7,
	author = {Farquhar, Sebastian and Kossen, Jannik and Kuhn, Lorenz and Gal, Yarin},
	title = {Detecting hallucinations in large language models using semantic entropy},
	journal = {Nature},
	volume = {630},
	number = {8017},
	pages = {625--630},
	year = {2024},
	doi = {10.1038/s41586-024-07421-0},
}

@article{bib14,
	title = {HaluGNN: Hallucination Detection in Large Language Models Using Graph Neural Network},
	journal = {Expert Systems with Applications},
	pages = {130857},
	year = {2025},
	issn = {0957-4174},
	doi = {10.1016/j.eswa.2025.130857},
	url = {https://www.sciencedirect.com/science/article/pii/S0957417425044720},
	author = {Linggang Kong and Yunlong Zhang and Xiaofeng Zhong and Haoran Fu and Yongjie Wang and Huijun Liu},
}

@article{bib16,
    title={CausalSR: Structural causal model-driven super-resolution with counterfactual inference},
    author={Zhengyang Lu and Bingjie Lu and Feng Wang},
    journal={Neurocomputing},
    issue={No.C},
    pages={130375},
    year={2025},
    doi = {10.1016/j.neucom.2025.130375}
}

@article{bib20,
    title={Multi-perspective consistency checking for large language model hallucination detection: a black-box zero-resource approach},
    author={Linggang Kong and Xiaofeng Zhong and Jie Chen and Haoran Fu and Yongjie Wang},
    journal={Front Inform Technol Electron Eng},
    year={2025},
    doi = {10.1631/FITEE.2500180}
}

@article{bib27,
    title={SCM-GNN: A Graph Neural Network-Based Multi-Antenna Spectrum Sensing in Cognitive Radio},
    author={Youqiang Dong and Min Zhang and Xi Cheng and Hai Wang},
    journal={IEEE Transactions on Cognitive Communications and Networking},
    issue={1},
    pages={127-144},
    year={2025},
    doi = {10.1109/TCCN.2024.3431923}
}

\appendix

\section{Details about Experimental Settings}
\label{appendixA}
\subsection{Details about Models}
\label{appendixA1}
Llama2-7B, Qwen2-7B and Mistral-7B are selected for their accessible internal latent spaces and the moderate dimensionality of the hidden states, facilitating both experimental implementation and validation. We employ the pre-trained parameters provided directly by the Hugging Face throughout our experiments. During inference, we adhere to the default generation configuration with the temperature set to 0.6, while top-k and top-p sampling set to 50 and 0.9, respectively.

\subsection{Details about Datasets}
\label{appendixA2}
To rigorously evaluate the detection performance of our CausalGaze framework across varied knowledge domains and complexities, we curate an experimental corpus from four distinct datasets.  Specifically, \textbf{TruthfulQA} includes 817 question-answering (QA) pairs with testing the model’s resilience against factually challenging questions \citep{bib14}; \textbf{TriviaQA} consists of 9960 deduplicated QA pairs for assessing performance on a broader and more conventional knowledge base; \textbf{SciQ} contains 1000 QA pairs which are used to probe the model’s capabilities in a specialized scientific domain; and \textbf{HaluEval} (QA subset) inclueds 10k QA pairs technically for hallucination detection tasks. To facilitate experimental implementation, we select 1000 samples from the aforementioned datasets, which are divided into 400 for training, 200 for validation, and the remaining for testing.

\subsection{Baselines Methods}
\label{appendixA3}
This section provides a brief introduction of the involved baseline methods for hallucination detection.

\textbf{Consistency-based approaches:} Consistency is a fundamental concept in logic, defined as the absence of contradictory statements within a system. To characterize the internal consistency of LLMs, consistency-based hallucination detection methods rely on multiple samplings, which facilitate zero-resource fact-checking and enable the verification of responses from arbitrary LLMs without the reliance on any external databases or evidence. \textbf{SelfCheckGPT} builds on a straightforward intuition: If LLMs possess knowledge of a given concept, the sampled responses tend to be similar and factually consistent. Conversely, in the case of hallucinated content, randomly sampled responses are likely to diverge and contradict one another. \textbf{EigenScore} operates on the similar premise but computes consistency scores using the LLM internal states, specifically by evaluating the similarity of latent spaces across multiple samples.

\textbf{Logit-based approaches:} For the traditional machine learning classification models, the maximum Softmax probability indicates the confidence level of the results and has been widely used as the metric for uncertainty evaluation. Extending it to the long-sequence token generation tasks of LLMs, \textbf{Perplexity} defines the uncertainty of the generated response as the joint probability of the constituent tokens:

\begin{equation}
	P(y|x,\Theta)=-\frac{1}{T}\sum_t\log{p(y_t|y_{<t},x)}
\end{equation}
where $x$ is the prompt, $T$ denotes the length of the sequence, and $p(y_t|y_{<t},x)$ represents the maximum Softmax probability of $t\mbox{-}th$ token. Given that the perplexity of shorter sequences generally exhibit lower, the joint probability is normalized by the output sequence length $T$.

However, considering that different tokens contribute unevenly to the sentence, the average probability fails to effectively capture the uncertainty. Multiple responses can be obtained during inference via the top-p or top-k sampling strategies. \textbf{LN-Entropy} measures the uncertainty as follows:

\begin{equation}
	H(\mathcal{Y}|x,\Theta) \\
	=-\mathbb{E}_{y\in\mathcal{Y}}\frac{1}{T_y}\sum_t\log{p(y_t|y_{<t}},x)
\end{equation}
where $\mathcal{Y}=[y^1,y^2,...,y^{K-1},y^K]$ denotes $K$ sampled responses.

The above methods assess uncertainty and entropy solely from a token-level perspective. However, measuring uncertainty in natural language is challenging due to semantic equivalence, where distinct sentences can convey identical meanings. \textbf{Semantic Entropy} addresses it by incorporating linguistic invariance arising from shared meaning. It employs semantic equivalence to cluster $K$ responses into $c$ classes, and then computes the Semantic Entropy as the entropy distribution over the semantic space:
\begin{equation}
	p(c|x)=\sum_{s\in c}p(s|x)=\sum_{s\in c}\prod_tp(s_t|s_{<t},x)
\end{equation}
\begin{equation}
	SE(x)=-\sum_cp(c|x)\log{p(c|x)}
\end{equation}

\textbf{Self-evaluation approach:} LLMs are also frequently utilized to assist in or directly judge the correctness of responses, known as LLM-as-a-Judge. \textbf{P(True)} performs fact-checking by querying LLMs with specific prompts and evaluating the probability that the response is correct. The prompt that we use for P(True) is as follows:
\\

\myaclprompt{
	\textbf{Provide the probability that the following answer for the question is correct. Give ONLY the probability value between 0.0 and 1.0, no other words or explanation.}
	\medskip

	\textbf{Question:} \{Question\}
	
	\textbf{Answer:} \{Answer\}
}

\textbf{Classification-based approaches:} The hallucination detectors are mainly trained on the latent space of LLMs. \textbf{SAPLMA} trains a classifier using the hidden states of specific layers in LLMs. \textbf{LLM-Check} uses the method of attention mechanism kernel similarity analysis to conduct hallucination detection. And it has proved that the differences in model sensitivity to hallucinated or truthful contents reflects in the rich semantic representations present both in hidden states and the pattern of attention maps. \textbf{ICR Probe} firstly quantifies the global contribution of modules to the hidden states’ updates of all layers, which are then used to train a hallucination probe. \textbf{HaluGNN} models the hidden states and attention maps as weighted directed graphs, and train a GNN-based classifier. However, the classification-based approaches scarcely concerns the interpretability of the detection model and the results.

\subsection{CausalGaze Training}
\label{appendixA4}
\textbf{Model Architecture.} The CasusalGaze model consists of five layers, including a projected layer, a refinement layer, two GAT layers and a linear layer. The dimension of each layer passes through $d \to 128 \to 128 \to 64 \to 64 \to 2$. Each hidden layer except the refinement layer employs the ReLU activation function and applies the dropout ($p=0.2$) to prevent overfitting.

\textbf{Details of Training.} 
We set the hyper-parameters $a=1$ and $b=0$ in Equation \ref{eq4}, and minum value of the $Clamp$ is $0$, and $\lambda$ is set to $0.02$ in Equation \ref{eq8}.
The loss function is the binary crossentropy loss with a regularization term that promotes sparsity and coherence in the causal structure. 
We use the AdamW optimizer and \texttt{CosineAnnealingWarmRestarts} learning rate scheduler, which implements cosine annealing of the learning rate with periodic warm restarts, to train the CausalGaze model. And the initial learning rate is $\num{1e-4}$ and the scheduler parameters are set as $T_0=10$, $T_{mult}=2$, $\eta_{min}=\num{1e-6}$. 
The model is trained for 50 epochs with the early stop mechanism ($patience=20$) and a batch size of 8. 
To obtain stable and reliable results, we perform multiple runs and take the average as the final results.

\begin{table}[t]
	\centering
	\resizebox{\columnwidth}{!}{
		\begin{tabular}{lcccc}
			\hline
			\textbf{Methods}    & \textbf{TruthfulQA}    & \textbf{TriviaQA}      & \textbf{SciQ}          & \textbf{HaluEval}      \\ \hline
			Lookback Lens       & 0.603 / 0.633          & 0.771 / 0.781          & 0.726 / 0.731          & 0.752 /0 .795          \\
			Factoscope          & 0.552 / 0.584          & 0.714 / 0.760          & 0.682 / 0.702          & 0.742 / 0.788          \\
			TAD                 & 0.804 / 0.832          & 0.844 / 0.849          & 0.815 / 0.833          & 0.874 / 0.856          \\
			\textbf{CausalGaze} & \textbf{0.881 / 0.852} & \textbf{0.861 / 0.895} & \textbf{0.839 / 0.855} & \textbf{0.928 / 0.938} \\ \hline
	\end{tabular}}
	\caption{ AUROC / PR-AUC Performance of different methods on Llama2-7B.}
	\label{add2}
\end{table}

\begin{table*}[t]
	\centering
	\resizebox{\textwidth}{!}{
		\begin{tabular}{ccccccccccc}
			\toprule
			\multirow{2}{*}{\textbf{LLMs}}      & \multirow{2}{*}{\textbf{Projection Layer}} & \multirow{2}{*}{\textbf{Dimension}} & \multicolumn{2}{c}{\textbf{TruthfulQA}} & \multicolumn{2}{c}{\textbf{TriviaQA}} & \multicolumn{2}{c}{\textbf{SciQ}} & \multicolumn{2}{c}{\textbf{Halueval}} \\ \cline{4-11} 
			&                                            &                                     & AUROC              & F1-Score           & AUROC             & F1-Score          & AUROC           & F1-Score        & AUROC             & F1-Score          \\ \midrule
			\multirow{5}{*}{\textbf{Qwen2-7B}}   & w/o                                        & 3584                                & 0.5925             & 0.5663             & 0.6043            & 0.5728            & 0.5964          & 0.6459          & 0.5620            & 0.6588            \\\cline{2-11}
			& \multirow{4}{*}{with}                      & 64                                  & 0.8330             & 0.7845             & 0.8764            & 0.7890            & 0.8976          & \textbf{0.8423} & 0.8946            & 0.7892            \\
			&                                            & 128                                 & \textbf{0.8680}    & \textbf{0.8071}    & \textbf{0.9106}   & \textbf{0.8280}   & \textbf{0.9328} & 0.8283          & \textbf{0.9220}   & \textbf{0.8360}   \\
			&                                            & 256                                 & 0.8357             & 0.7786             & 0.8873            & 0.8130            & 0.8741          & 0.7740          & 0.8863            & 0.7538            \\
			&                                            & 512                                 & 0.8588             & 0.7943             & 0.8743            & 0.7924            & 0.8661          & 0.7548          & 0.8750            & 0.7630            \\ \midrule
			\multirow{5}{*}{\textbf{Mistral-7B}} & w/o                                        & 4096                                & 0.6146             & 0.6273             & 0.5738            & 0.5460            & 0.5964          & 0.6248          & 0.5692            & 0.6469            \\\cline{2-11}
			& \multirow{4}{*}{with}                      & 64                                  & 0.8560             & 0.7533             & 0.8690            & 0.7793            & 0.8452          & {0.7726} & 0.8824            & 0.6926            \\
			&                                            & 128                                 & \textbf{0.8851}    & \textbf{0.7971}             & \textbf{0.8880}   & \textbf{0.8113}   & \textbf{0.8550} & \textbf{0.7860}          & \textbf{0.9127}   & \textbf{0.8292}   \\
			&                                            & 256                                 & 0.8663             & {0.7894}    & 0.8770            & 0.7957            & 0.8467          & 0.7460          & 0.8840            & 0.7850            \\
			&                                            & 512                                 & 0.8580             & 0.7587             & 0.8659            & 0.7893            & 0.8198          & 0.7319          & 0.8649            & 0.7748            \\ \bottomrule
	\end{tabular}}
	\caption{The contribution of the Projection Layer and its dimentions on Qwen2-7B and Mistral-7B across different datasets. The best results are highlighted in bold.}
	\label{tab3}
\end{table*}

% Please add the following required packages to your document preamble:
% \usepackage{multirow}
\begin{table*}[t]
	\centering
	\resizebox{\textwidth}{!}{
		\begin{tabular}{cccccccccc}
			\hline
			\multirow{2}{*}{\textbf{LLMs}}       & \multirow{2}{*}{\textbf{Layer Number}} & \multicolumn{2}{c}{\textbf{TruthfulQA}} & \multicolumn{2}{c}{\textbf{TriviaQA}} & \multicolumn{2}{c}{\textbf{SciQ}} & \multicolumn{2}{c}{\textbf{Halueval}} \\ \cline{3-10} 
			&                                        & AUROC              & F1-Score           & AUROC             & F1-Score          & AUROC           & F1-Score        & AUROC             & F1-Score          \\ \hline
			\multirow{5}{*}{\textbf{Qwen2-7B}}   & 16                                     & 0.8718             & 0.7836             & 0.9005            & 0.8194            & 0.9114          & 0.8445          & 0.9260            & 0.8004            \\
			& 20                                     & 0.8776             & 0.7893             & \textbf{0.9125}   & \textbf{0.8304}   & 0.9268          & \textbf{0.8675} & \textbf{0.9335}   & \textbf{0.8421}   \\
			& 24                                     & \textbf{0.8892}    & 0.7747             & 0.8994            & 0.7961            & 0.8993          & 0.8593          & 0.9154            & 0.8266            \\
			& 28                                     & 0.8757             & 0.7694             & 0.8873            & 0.7868            & 0.8858          & 0.8495          & 0.9255            & 0.8273            \\
			& 32                                     & 0.8680             & \textbf{0.8071}    & 0.9106            & 0.8280            & \textbf{0.9328} & 0.8283          & 0.9220            & 0.8360            \\ \hline
			\multirow{5}{*}{\textbf{Mistral-7B}} & 16                                     & 0.8673             & 0.8057             & 0.8861            & 0.8090            & 0.8479          & 0.7505          & 0.8889            & 0.8162            \\
			& 20                                     & 0.8862             & \textbf{0.8203}    & \textbf{0.8882}   & {0.7806}   & 0.8361          & 0.7260          & 0.9073            & \textbf{0.8347}   \\
			& 24                                     & \textbf{0.8990}    & 0.8183             & 0.8637            & 0.7962            & \textbf{0.8693} & 0.7348          & \textbf{0.9173}   & 0.7963            \\
			& 28                                     & 0.8735             & 0.7769             & 0.8643            & 0.7981            & 0.8463          & 0.7450          & 0.9065            & 0.7990            \\
			& 32                                     & 0.8851             & 0.7971             & 0.8880            & \textbf{0.8113}            & 0.8550          & \textbf{0.7860} & 0.9127            & 0.8292            \\ \hline
	\end{tabular}}
	\caption{The impact of causal graphs from different layers of Qwen2-7B and Mistral-7B on the detection performance.}
	\label{tab4}
\end{table*}

\section{Details about Experiment}
\label{appendixB}

\subsection{Main Results}
\label{appendixBadd}

We further evaluate our method CausalGaze against recent baselines (Lookback Lens \cite{Lookback}, Factoscope \cite{Factoscope}, TAD \cite{TAD}) and incorporate a new metric PR-AUC for evaluation. The experiments are implemented on Llama2-7B, as shown in Table \ref{add2}.
As demonstrated in the results, CausalGaze consistently outperforms all baselines. We attribute it to several key factors: (1) Lookback Lens is designed for contextual hallucinations. While the datasets used in our work are with no external context, its mechanism reduces to a simple similarity measure, failing to capture internal veracity; (2) Factoscope and TAD rely on token probabilities and attention maps. Being sensitive to local fluctuations and specific entities, they struggle with complex logical hallucinations; (3) CausalGaze captures global causal dependencies by modeling the full question-answering (QA) as a causal graph. From the information-theoretic perspective, CausalGaze processes a richer set of signals, leading to superior stability and accuracy.

\begin{figure}[t]
	\centering 
	\includegraphics[width=\columnwidth]{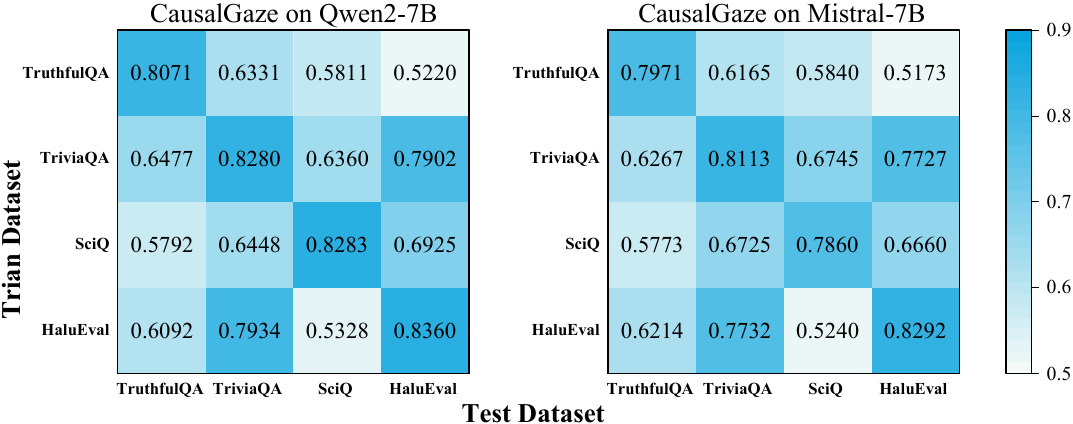}
	\caption{Cross-dataset generalization analysis for CausalGaze on Qwen2-7B and Mistral-7B. Each subplot displays the F1-Score when the model is trained on the row dataset and tested on the column dataset, with values annotated in each cell.}
	\label{cross_appendix}
\end{figure}

\subsection{Generalization}
\label{appendixB1}
In Section~\ref{subsec5.2}, we demonstrate the generalization ability of the proposed CausalGaze on Llama2-7B. In this section, we provide additional generalization results on Qwen2-7B and Mistral-7B as shown in Figure~\ref{cross_appendix}, which show the effectively robust generalization ability of our approach.

\subsection{Contribution of the Projection Layer and its Dimension}
\label{appendixB2}
In Section~\ref{subsec5.3}, we demonstrate the contribution of the projection layer and its dimension on the detection performance for Llama2-7B. In this section, we provide additional experimental results on Qwen2-7B and Mistral-7B as shown in Table~\ref{tab3}. The results consistently indicate that the projection layer is of considerable interest for the detection performance.

% Please add the following required packages to your document preamble:
% \usepackage{multirow}

\subsection{Layer Number}
\label{appendixB3}
In Section~\ref{subsec5.3}, we demonstrate the impact of dynamic causal graphs from different layer of  Llama2-7B on the detection performance. In this section, we provide additional experimental results on Qwen2-7B and Mistral-7B as shown in Table~\ref{tab4}.

\subsection{Gradient-guided Information}
\label{appendixB4}
Without gradients, the model reverts to the raw graph that treats all attention edges with equal importance, failing to disentangle the causal reasoning paths from noisy and spurious correlations. Random gradients lead to the misallocation of attention in the detection model between spurious edges and genuinely contributory edges. Raw attention weights $A$ only reflect the co-occurrence patterns during the LLM's forward infernece. As documented in prior literature, high attention weights often focus on spurious correlations (e.g., stop words or position bias) that do not necessarily drive the final decision.
While the learned gradients $S$ assigns higher importance to edges that have a larger influence on the veracity prediction, allowing the gating mechanism to suppress those shortcut edges that do not contribute to the factual reasoning.

\begin{table}[t]
	\centering
	\resizebox{\columnwidth}{!}{
		\begin{tabular}{lccc}
			\hline
			\textbf{Method} & \multicolumn{1}{l}{\textbf{Input memory}} & \multicolumn{1}{l}{\textbf{Model parameters}} & \multicolumn{1}{l}{\textbf{Inference latency}} \\ \hline
			SAPLMA          & 608KB                                     & 109K                                          & $\sim$0.307ms                                  \\
			HaluGNN         & 614KB                                     & 213K                                          & $\sim$0.903ms                                  \\
			SelfCheckGPT    & 608KB                                     & -                                             & $\sim$10.32s                                   \\
			CausalGaze      & 614KB                                     & 575K                                          & $\sim$1.731ms                                  \\ \hline
	\end{tabular}}
	\caption{Computation overhead comparison across different baselines.}
	\label{tab5}
\end{table}

\subsection{Computation Overhead}
\label{appendixB5}
We also benchmark CausalGaze against three representative baselines, including SAPLMA, HaluGNN and SelfCheckGPT, in terms of memory footprint (e.g., input and model) and inference latency to provide memory and inference overhead comparison. The experiments are conducted on a single NVIDIA A800 (80GB) GPU with Llama2-7B as the backboneone, and the example is from TriviaQA with input sequence length $L=38$, feature dimension $D=4096$. The results on Llama2-7B are summarized in Table~\ref{tab5}. 

While CausalGaze exhibits a higher model memory and latency compared to other methods, its absolute overhead remains extremely low and highly practical for real-world deployment. The inference latency is approximately 1.731 ms, which is negligible compared to the time required for an LLM to generate a single token. This ensures that our detection framework can operate as a real-time safety monitor without bottlenecking the generation process. The slight increase in complexity of CausalGaze (about +0.8 ms) is a deliberate trade-off to enable active causal intervention. This process allows us to filter out incidental noise and achieve competitive detection performance (e.g., +6.1\% AUROC improvement over SAPLMA) with interpretability.

\end{document}